\documentclass{article} 
\usepackage[preprint]{colm2026_conference}

\usepackage{microtype}
\usepackage{hyperref}
\usepackage{url}
\usepackage{graphicx}
\usepackage{amsmath}
\usepackage{amssymb}
\usepackage{mathtools}
\usepackage{amsthm}
\usepackage{bbm}
\usepackage{booktabs,multirow,tabularx,array}
\usepackage{makecell}
\usepackage{listings}
\usepackage{xcolor}
\usepackage{algorithmic}
\usepackage{algorithm}
\usepackage{enumitem}
\usepackage{titlesec}
\usepackage{parskip}
\usepackage{fancyhdr}
\usepackage{subcaption} 

\usepackage{wrapfig}
\usepackage[dvipsnames]{xcolor}
\usepackage[most]{tcolorbox}

\usepackage{lineno}

\definecolor{darkblue}{rgb}{0, 0, 0.5}
\hypersetup{colorlinks=true, citecolor=darkblue, linkcolor=darkblue, urlcolor=darkblue}

\newcommand{\modelname}{ThinkTwice}
\newcommand{\xhdr}[1]{\vspace{2mm}\noindent{{\bf #1.}}}

\definecolor{AccentBlue}{HTML}{264E86}
\definecolor{SoftBlue}{HTML}{B0CCEC}
\definecolor{SoftGray}{HTML}{F6F7F9}
\definecolor{SoftGreen}{HTML}{EAF7EF}
\definecolor{SoftRed}{HTML}{FCEDEE}
\definecolor{RuleGray}{HTML}{D7DEE8}

\newcommand{\correct}{\textcolor{ForestGreen}{\textsc{correct}}}
\newcommand{\wrong}{\textcolor{BrickRed}{\textsc{wrong}}}

\newtcolorbox{summarybox}{
  enhanced,
  breakable,
  colback=SoftBlue,
  colframe=AccentBlue!65,
  boxrule=0.7pt,
  arc=2pt,
  left=8pt,right=8pt,top=8pt,bottom=8pt,
  title={Reading guide},
  fonttitle=\bfseries,
}

\newtcolorbox{problembox}{
  enhanced,
  breakable,
  colback=white,
  colframe=RuleGray,
  boxrule=0.7pt,
  arc=2pt,
  left=8pt,right=8pt,top=8pt,bottom=8pt,
  title={Problem and role in the appendix},
  fonttitle=\bfseries,
}

\newtcolorbox{excerptbox}[1]{
  enhanced,
  breakable,
  colback=SoftGray,
  colframe=RuleGray,
  boxrule=0.6pt,
  arc=2pt,
  left=8pt,right=8pt,top=6pt,bottom=6pt,
  title={#1},
  fonttitle=\bfseries,
}

\title{ThinkTwice: Jointly Optimizing Large Language Models for Reasoning and Self-Refinement}

\author{Difan Jiao$^{1\dagger}$, Qianfeng Wen$^{1\dagger}$, Blair Yang$^{1,2}$, Zhenwei Tang$^{1}$ \& Ashton Anderson$^{1}$ \\
$^{1}$Department of Computer Science, University of Toronto \quad $^{2}$Coolwei AI Lab \\
$^\dagger$Equal contribution. Contact: \texttt{\{difanjiao, ashton\}@cs.toronto.edu}
}

\begin{document}

\ifcolmsubmission
\linenumbers
\fi

\maketitle

\begin{abstract}


We introduce \modelname{}, a simple two-phase framework that jointly optimizes LLMs to solve reasoning problems and refine the answers, based on Group Relative Policy Optimization (GRPO). In each pair of training steps, \modelname{} first optimizes the model on solving reasoning problems, then optimizes it on refining its own solutions to the same problems, using the same binary correctness reward in both phases without correctness signals or critique annotations. Across five mathematical reasoning benchmarks and two
model families including Qwen3-4B and Olmo3-7B, \modelname{} substantially improves both reasoning and refinement performance over competitive online policy optimization baselines. Specifically, on Qwen3-4B, \modelname{} outperforms GRPO on AIME by 5 percentage points before refinement and by 11.5 points after one self-refinement step, measured by pass@4. Analysis of the training dynamics of \modelname{} reveals an implicit \emph{rectify-then-fortify} curriculum: refinement predominantly corrects errors early in training and naturally shifts toward preserving already-correct solutions as the model improves, yielding a more rectified reward signal. Our work establishes joint training of reasoning and self-refinement as a principled and effective methodology for RLVR. Our codebase is available at \url{https://github.com/CSSLab/ThinkTwice}.

\end{abstract}
\section{Introduction}

Reinforcement learning with verifiable rewards (RLVR) has emerged as
an effective paradigm for improving the reasoning capabilities of
large language models (LLM)~\citep{shao2024deepseekmath, guo2025deepseek}. Yet even strong reasoners can produce solutions that contain correctable errors, such as incomplete derivations, algebraic mistakes, or unproductive solution paths. A natural strategy from human problem-solving is \textit{self-refinement}: people routinely revisit initial solutions, identify errors, and revise their reasoning~\citep{polya1945solve}. There is growing evidence that self-refinement can similarly benefit LLMs to recover on challenging problems where initial attempts narrowly fail~\citep{gou2023critic, weng2023large}. 

\begin{figure*}[t]
    \centering
    \includegraphics[width=1.0\linewidth]{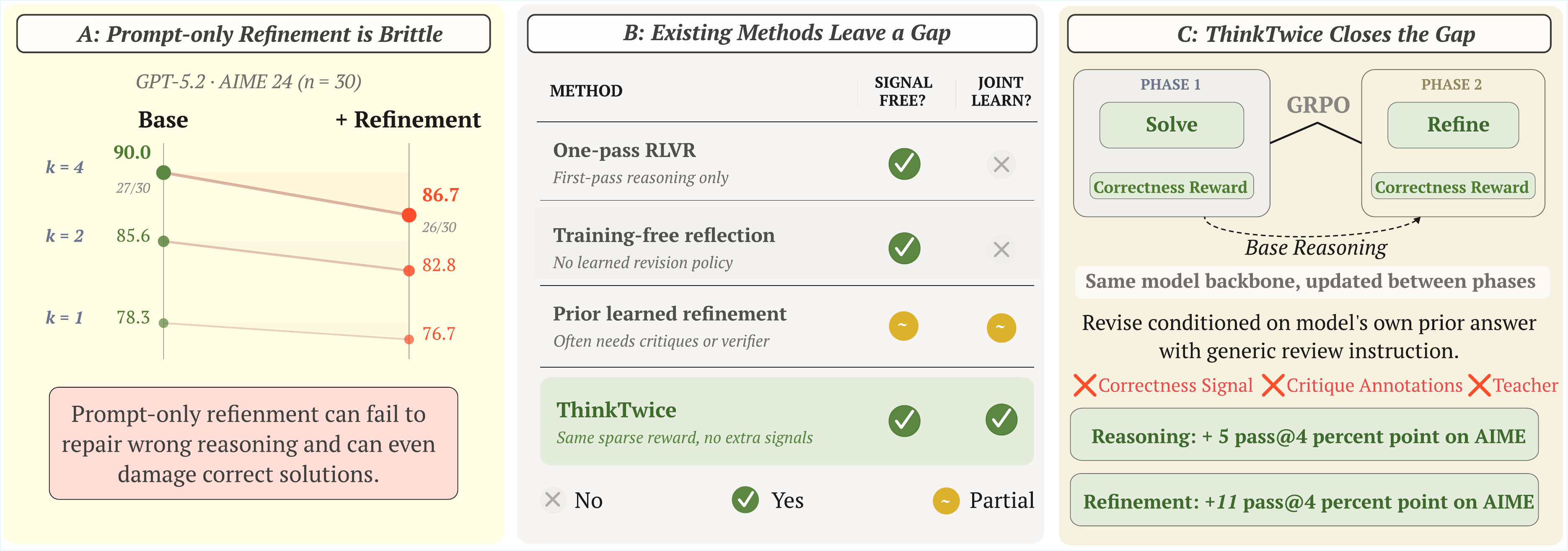}
    \caption{(A) Prompt-only reflection can reduce top frontier LLM's performance on AIME24, indicating brittleness. (B) \modelname{} compared with existing method families. (C) ThinkTwice addresses these gaps by sequentially training a shared model backbone—first solving, then reflecting—yielding significant gains (+5 points reasoning, +11 points refinement) on AIME with Qwen3-4B.}
    \label{fig:ThinkTwice}
\end{figure*}

However, existing approaches to improving self-refinement in LLMs,
broadly categorized into training-free and training-based methods,
each have notable limitations. First, training-free methods~\citep{madaan2023self, shinn2023reflexion} prompt the model to critique and revise at inference time but do not learn a reusable refinement policy. Moreover, prompt-only refinement remains
brittle even for frontier models. As illustrated in Figure~\ref{fig:ThinkTwice}(A), we observe a slight performance \emph{decrease} from the frontier LLM on AIME24 after self-refinement prompting. Second, although training-based methods learn refinement behavior, existing methodologies typically rely on process
supervision~\citep{Zhang2024ReSTMCTS, cui2025process}, critique annotations~\citep{kumar2024training, zhang2025critique}, or explicit signals indicating whether the initial answer is correct~\citep{ma2025s2r}. Such supervision is not always available in practice: at the frontier, no stronger model exists to provide reliable critique, and human oversight may be insufficient for challenging problems~\citep{burns2023weak}. This limits existing training-based approaches to settings with rich external feedback.

In this work, we introduce \modelname{}, a unified RLVR framework
that jointly optimizes reasoning and self-refinement without access to external information. In each pair
of training steps, \modelname{} first optimizes the model on solving
a batch of reasoning problems, then optimizes it on refining its own
solutions to the same problems (``thinking twice''), using a generic review instruction
and the same binary correctness reward in both phases. The
refinement phase gives the model a second attempt at each problem conditioned on its prior solution, allowing it to recover from
correctable errors that the first attempt missed. As shown in
Figure~\ref{fig:ThinkTwice}(B, C), \modelname{} serves as a simple, effective method that doesn't require correctness signals, critique annotations, or external teacher models.

We train \modelname{} with Qwen3-4B and Olmo3-7B and evaluate on a suite of five mathematical reasoning benchmarks.  \modelname{} substantially improves both reasoning and self-refinement capabilities over competitive baselines. Specifically, on Qwen3-4B, \modelname{} outperforms GRPO on the challenging AIME problems by 5 percentage points before refinement and by 11.5 points after one step of self-refinement, measured by pass@4. Across both models, \modelname{} achieves the highest average performance for both reasoning and self-refinement, and when used as a refiner applied to other baselines' solutions, \modelname{} also achieves the highest score.

To understand how \modelname{} works, we analyze its training
dynamics and find that the refinement phase gives rise to an implicit
\textit{rectify-then-fortify} curriculum: early in training, refinement
predominantly corrects errors in failed solutions, and as the
model improves, it naturally shifts toward preserving and polishing
already-correct ones. This dynamic yields a more rectified reward
signal, as the refinement phase recovers useful learning signal from
problems where the base attempt alone falls short. Additionally, \modelname{} adds only 3\% training overhead per step, and it reaches its best checkpoint in 16\% less wall-clock time than GRPO, as the enriched training signal empirically leads to faster convergence.


\section{Related Work}
\label{sec:related_work}

\paragraph{Reinforcement learning with verifiable rewards (RLVR).}
Recent progress in reasoning post-training has been driven by reinforcement learning with verifiable rewards (RLVR), where models are optimized using automatically checkable outcome signals such as exact answers. DeepSeekMath introduced Group Relative Policy Optimization (GRPO), a practical critic-free PPO-style recipe for this setting, and DeepSeek-R1 further showed that large-scale outcome-only RL can elicit strong reasoning behavior without process supervision \citep{shao2024deepseekmath,guo2025deepseek}. A large follow-up literature has since expanded the RLVR design space. Open reproductions and training recipes such as Open-Reasoner-Zero, SimpleRL-Zoo, and Skywork-OR1 helped make large-scale RLVR more reproducible, while systems work such as AReaL improved training efficiency at scale \citep{hu2025open,zeng2025simplerl,he2025skywork,fu2025areal}. On the optimization side, DAPO improves long-CoT training stability; Dr.~GRPO analyzes optimization bias in GRPO; GSPO moves from token-level to sequence-level importance ratios and clipping; and newer variants such as GMPO, GPG, and shrinkage baselines revisit ratio aggregation, simplification, response-length bias, and baseline variance \citep{yu2025dapo,liu2025understanding,zheng2025group,zhao2025geometric,chu2025gpg,zeng2025shrinking}.

\paragraph{Self-refinement.}
Self-refinement aims to improve an initial response by generating feedback or verification and then revising the answer \citep{madaan2023self}. A first class of methods is training-free and operates purely at inference time, such as Self-Refine, Reflexion, self-verification prompting, CRITIC, and self-consistency decoding \citep{madaan2023self,shinn2023reflexion,weng2023large,gou2023critic,wang2022selfconsistency}. These approaches show that extra test-time computation, self-feedback, or tool-based critique can improve outputs, but they do not learn a reusable refinement policy; moreover, prompt-only self-correction can be unreliable for reasoning without external feedback \citep{huang2023large}. A second class of work trains refinement behavior more directly, including methods based on verifiers \citep{cobbe2021training,zhang2024small}, process supervision \citep{uesato2022solving,lightman2023let,wang2024math,yuan2024free,cui2025process}, critique data \citep{xi2024enhancing,yu2025training}, synthetic correction traces \citep{welleck2022generating,qu2024recursive,xiong2025self,zhao2025boosting}, or multi-turn RL with explicit self-verification or critique objectives \citep{kumar2024training,ma2025s2r,he2025rise,jiang2025pag,zhang2025incentivizing,zhang2025critique}. \modelname{} is closest to the RL-based self-refinement line, but is deliberately simpler: it uses a shared policy, a generic review instruction, and the same final-answer correctness reward in both phases, without process labels, critique annotations, correctness hints, or an explicit verifier indicating whether the initial answer is correct. A more detailed comparison is provided in Appendix~\ref{app:related-work-table-sec}.

\section{Methodology}

We introduce \modelname{}, an RLVR framework that jointly optimizes reasoning and self-refinement without verification---i.e., without access to external signals indicating whether the intermediate steps or solutions are correct or not. We briefly recap GRPO in Section~\ref{sec:prelim}, then describe how we formulate the self-refinement problem in Section~\ref{sec:method_refine}, and finally explain how we integrate both reasoning and self-refinement optimization in Section~\ref{sec:thinktwice}.

\begin{figure}
    \centering
    \includegraphics[width=0.9\linewidth]{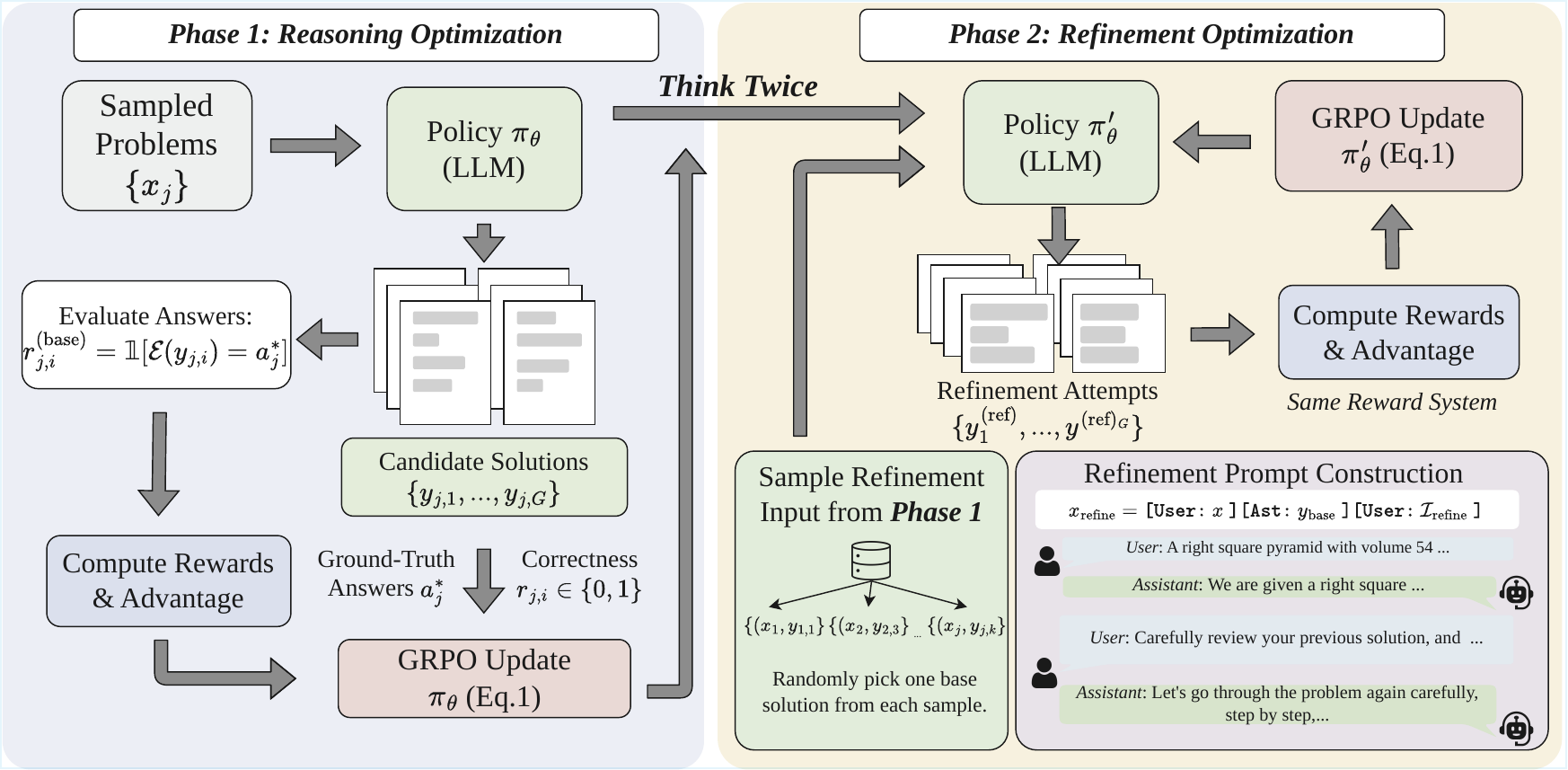}
    \caption{\modelname{} at a glance.}
    \label{fig:OurMethod}
\end{figure}

\subsection{Preliminaries}
\label{sec:prelim}

We use Group Relative Policy Optimization (GRPO)~\citep{shao2024deepseekmath, guo2025deepseek} as our backbone reinforcement learning (RL) algorithm. GRPO eliminates the need for a separate critic model by computing advantages from grouped samples, which offers better stability and efficiency during LLM RL.

Let $\pi_\theta$ denote the language model policy parameterized by $\theta$. For a given input $x$, GRPO samples a group of $G$ responses $\{y_1, \ldots, y_G\} \sim \pi_{\theta_{\text{old}}}(\cdot | x)$, where each generation of $y_i$ is often called a rollout, and optimizes:

\begin{equation}
\mathcal{J}_{\text{GRPO}}(\theta) = \mathbb{E}_{x, \{y_i\}_{i=1}^G} \left[ \frac{1}{G} \sum_{i=1}^{G} \mathcal{L}_i - \beta \mathbb{D}_{\text{KL}}(\pi_\theta \| \pi_{\text{ref}}) \right],
\label{eq:grpo}
\end{equation}

where $\mathcal{L}_i = \min\left( \rho_i A_i, \operatorname{clip}(\rho_i, 1{-}\epsilon, 1{+}\epsilon) A_i \right)$ denotes the clipped surrogate objective, $\rho_i = \pi_\theta(y_i|x) / \pi_{\theta_{\text{old}}}(y_i|x)$ is the importance ratio, $\epsilon$ is the clipping threshold, and $\beta$ controls the KL penalty against the reference policy $\pi_{\text{ref}}$.

The advantage $A_i$ is computed via group normalization:
\begin{equation}
A_i = \frac{r_i - \text{mean}(r_1, \ldots, r_G)}{\text{std}(r_1, \ldots, r_G)},
\label{eq:advantage}
\end{equation}
where $r_i$ is the reward for response $y_i$. For mathematical reasoning tasks with verifiable answers, we use outcome-based correctness rewards: $r_i = \mathbbm{1}[\mathcal{E}(y_i) = a^*]$, where $\mathcal{E}$ extracts the answer from $y_i$ and $a^*$ is the ground-truth answer.

\subsection{Self-Refinement Formulation}
\label{sec:method_refine}

In our self-refinement framework, we use \textit{base solution} to refer to an initial solution attempt that the model will subsequently self-refine into a \textit{refined solution}. Given a problem $x$ and a base solution $y_{\text{base}}$ generated by the current policy, we construct the refinement prompt as a multi-turn conversation:

\begin{equation}
x_{\text{refine}} = \texttt{[User:}~x~\texttt{][Ast:}~y_{\text{base}}~\texttt{][User:}~\mathcal{I}_{\text{refine}}~\texttt{]},
\label{eq:refine_prompt}
\end{equation}

where $\mathcal{I}_{\text{refine}}$ is a task-agnostic instruction that asks the model to review and improve its previous solution. Crucially, $\mathcal{I}_{\text{refine}}$ contains no indication of whether $y_{\text{base}}$ is correct or incorrect, distinguishing our approach from methods that rely on external critique signals. We use a general refinement instruction for \modelname{} as shown in Appendix~\ref{app:imp_details}.

\subsection{\modelname{}}
\label{sec:thinktwice}
\modelname{} alternates between two training phases within a unified GRPO framework, as depicted in Figure~\ref{fig:OurMethod}. In each pair of training steps, the model first optimizes on reasoning for a batch of problems, then optimizes on refining its own solutions to the same problems. This exposes the model to each problem twice per training batch under complementary objectives: solving from scratch and improving upon its own prior attempt.

\xhdr{Phase 1: Reasoning Optimization} For each problem $x$ sampled from the training set $\mathcal{D}$, we generate $G$ candidate solutions $\{y_1, \ldots, y_G\}$ using the current policy $\pi_\theta$. Each solution is evaluated with the correctness reward $r^{(\text{base})}_i = \mathbbm{1}[\mathcal{E}(y_i) = a^*]$, and the policy is updated via Eq.~\eqref{eq:grpo}, yielding $\pi_\theta'$. From the $G$ rollouts per problem $x$, we randomly select one base solution for the subsequent refinement phase.

\xhdr{Phase 2: Refinement Optimization} Using the selected base solutions from Phase 1, we construct refinement prompts following Eq.~\eqref{eq:refine_prompt}. The model with updated policy $\pi_\theta'$ then generates $G$ refinement attempts $\{y^{(\text{ref})}_1, \ldots, y^{(\text{ref})}_G\}$ for each refinement prompt. Since $\pi_\theta'$ has already been updated using reward signal from $x$ during Phase 1, it is better equipped to improve upon the base solutions produced by $\pi_\theta$. Each refined solution is evaluated using the same correctness reward $r^{(\text{ref})}_i = \mathbbm{1}[\mathcal{E}(y^{(\text{ref})}_i) = a^*]$, and the policy is updated via Eq.~\eqref{eq:grpo} as well, yielding $\pi_\theta''$.

We now elaborate on two key design choices in the refinement optimization phase.

\xhdr{Base Solution Sampling} Since GRPO relies on multiple rollouts for advantage estimation, we accordingly have multiple base solutions as candidates for $y_{\text{base}}$. In \modelname{}, we adopt the simplest sampling strategy: randomly picking one base solution per problem $x$. This design not only inherently covers the full spectrum of training samples, but also aims to establish an \emph{emergent curriculum}: early in training, when the model frequently produces incorrect solutions, the refinement phase predominantly trains on error correction. As reasoning accuracy improves, fewer incorrect solutions are available, and the refinement phase naturally shifts toward polishing already-correct solutions. This adaptive curriculum naturally steers training toward appropriately challenging refinement scenarios aligned with the model's current capability boundaries.

\xhdr{Refined Solution Reward} While $\mathcal{I}_{\text{refine}}$ provides high-level guidance for the refinement task, we do not use extra reward signals to enforce adherence to its prescribed structure---through RL, the model is free to develop its own refinement strategies. \modelname{} solely employs the binary signal that depends on the correctness of the refined solution. This reward signal encourages the model to: \textbf{(1)} detect and correct errors when the base solution is wrong, or \textbf{(2)} preserve and polish when the base solution is already correct. Because the refinement is formulated as a standard multi-turn conversation, the optimization process is seamlessly handled by native GRPO.

The training procedure is detailed in Algorithm~\ref{alg:OurMethod} and Appendix~\ref{app:imp_details}.

\begin{algorithm}[t]
\caption{ThinkTwice: Jointly Optimizing Reasoning and Refinement}
\label{alg:OurMethod}
\begin{algorithmic}[1]
\REQUIRE Training dataset $\mathcal{D}$, current policy $\pi_\theta$, training batch size $B$, group size $G$, refinement instruction $\mathcal{I}_{\text{refine}}$
\FOR{each training iteration $t = 1, 2, \ldots, T$}
    \STATE \textcolor{gray}{// \textbf{Phase 1: Reasoning}}
    \STATE Sample batch of problems $\{x_j\}_{j=1}^B$ from $\mathcal{D}$
    \FOR{each problem $x_j$}
        \STATE Generate solutions $\{y_{j,1}, \ldots, y_{j,G}\} \sim \pi_\theta(\cdot | x_j)$
        \STATE Compute rewards $r_{j,i} = \mathbbm{1}[\mathcal{E}(y_{j,i}) = a^*_j]$ for $i = 1, \ldots, G$
    \ENDFOR
    \STATE Update $\pi_\theta$ to $\pi_\theta'$ on reasoning samples
    \STATE Sample refinement pairs from: $\{(x_j, y_{j,i})\}_{j,i}$
    \STATE \textcolor{gray}{// \textbf{Phase 2: Refinement}}
    \FOR{each $(x, y_{\text{base}})$ in sampled pairs}
        \STATE Construct $x_{\text{refine}}$ via Eq.~\eqref{eq:refine_prompt}
        \STATE Generate refinements $\{y^{(\text{ref})}_1, \ldots, y^{(\text{ref})}_G\} \sim \pi_\theta'(\cdot | x_{\text{refine}})$
        \STATE Compute rewards $r^{(\text{ref})}_i = \mathbbm{1}[\mathcal{E}(y^{(\text{ref})}_i) = a^*]$ for $i = 1, \ldots, G$
    \ENDFOR
    \STATE Update $\pi_\theta'$ to $\pi_\theta''$ on refinement samples
\ENDFOR
\end{algorithmic}
\end{algorithm}

\section{Results}

\subsection{Experimental Setup}

We follow the training and evaluation protocol of Dr. GRPO~\citep{liu2025understanding}.
We train on math questions from the MATH training dataset~\citep{hendrycks2021measuring}
and evaluate on a suite of five benchmarks: AIME,
AMC, MATH500, Minerva Math~\citep{lewkowycz2022solving}, and
OlympiadBench~\citep{he2024olympiadbench}. All methods use the binary correctness reward via Math-Verify,\footnote{\url{https://github.com/huggingface/Math-Verify}}
with exact-match verification against ground-truth answers.

We experiment with two instruction-tuned models, Qwen3-4B-Instruct-2507~\citep{yang2025qwen3}
and OLMo3-7B-Instruct~\citep{olmo2025olmo}, as the refinement phase
requires the ability to follow multi-turn instructions. Our baselines include standard GRPO~\citep{shao2024deepseekmath}, Dr.~GRPO~\citep{liu2025understanding},
and DAPO~\citep{yu2025dapo}. We also include two training-free baselines, Reflexion~\citep{shinn2023reflexion} and Self-Refine~\citep{madaan2023self}, for refinement evaluation. Our experiments are implemented using
VERL~\citep{sheng2024hybridflow},\footnote{\url{https://github.com/verl-project/verl}}
with implementation details and hyperparameter configurations provided in Appendix~\ref{app:imp_details}.

Our evaluations are two-fold. In Section~\ref{sec:reasoning}, we test the reasoning capability of \modelname{} and baselines with direct prompting as a single-turn generation. In Section~\ref{sec:refinement}, we evaluate the
refinement capability using the multi-turn chat format, testing models on refining both their own generations and also those produced by others.

\subsection{Reasoning}
\label{sec:reasoning}

\begin{table*}[t]
\centering
\caption{Reasoning performance (pass@4, $\uparrow$) across five mathematical reasoning benchmarks. \textbf{Bold} and \underline{underline} denote the best and second-best results per dataset within each model.}
\vspace{0.2cm}
\label{tab:reasoning_results}
\begin{tabular}{l|ccccc|c}
\toprule
\textbf{Method} & \textbf{AIME} & \textbf{AMC} & \textbf{MATH500} & \textbf{Minerva} & \textbf{OlympiadBench} & \textbf{Average} \\
\midrule
\multicolumn{7}{l}{\textbf{Qwen3-4B}} \\
\quad Base Model & 29.18 & 64.87 & 88.47 & 39.61 & 57.90 & 56.01 \\
\quad GRPO & 39.06 & 75.36 & 91.86 & 41.03 & 63.80 & 62.22 \\
\quad DrGRPO & 35.46 & 77.65 & 91.83 & \underline{42.75} & 66.51 & 62.84 \\
\quad DAPO & \underline{42.54} & \textbf{80.68} & \underline{93.55} & 38.38 & \underline{67.50} & \underline{64.53} \\
\quad \textbf{\modelname{}} & \textbf{44.11} & \underline{79.59} & \textbf{93.60} & \textbf{42.94} & \textbf{67.60} & \textbf{65.57} \\
\midrule
\multicolumn{7}{l}{\textbf{OLMo3-7B}} \\
\quad Base Model & 32.81 & 68.77 & 89.87 & 40.63 & 61.36 & 58.69 \\
\quad GRPO & \textbf{39.38} & \underline{77.05} & \underline{92.28} & 41.13 & 62.42 & \underline{62.45} \\
\quad DrGRPO & 36.09 & 74.33 & 91.65 & 42.07 & \underline{65.09}  & 61.85 \\
\quad DAPO & 36.72 & 76.16 & 91.56 & \underline{42.39} & 63.80 & 62.12 \\
\quad \textbf{\modelname{}} & \underline{39.24} & \textbf{79.89} & \textbf{92.74} & \textbf{43.43} & \textbf{65.81} & \textbf{64.22} \\
\bottomrule
\end{tabular}
\end{table*}

For each benchmark problem, we sample $n{=}32$ independent solutions for Qwen3-4B and OLMo3-7B. Following~\citet{chen2021evaluating}, we report pass@4, the estimated probability that at least one of 4 sampled solutions is correct. The full reasoning pass@$k$ curves are provided in Figure~\ref{fig:passatk_base} at Appendix~\ref{app:additional}.

Table~\ref{tab:reasoning_results} reports the reasoning performance on both models. For Qwen3-4B, \modelname{} achieves the highest average score of 65.57\%, outperforming all competitive baselines. The gains are most pronounced on the most challenging AIME benchmark, where \modelname{} reaches 44.11\% compared to 39.06\% for GRPO. Across the remaining benchmarks, \modelname{} is consistently the best or second-best method. For OLMo3-7B, the overall trend mirrors that of Qwen3-4B. with \modelname{} achieving the highest average score. Notably, these improvements are obtained on direct prompting alone, before any self-refinement step is applied, which indicates that the refinement training phase in \modelname{} itself strengthens the model's reasoning capability.

\subsection{Refinement}
\label{sec:refinement}

\begin{table*}[t]
\centering
\caption{Self-refinement performance (pass@4, $\uparrow$) across five mathematical reasoning benchmarks. \textbf{Bold} and \underline{underline} denote the best and second-best results per dataset within each model. $\dagger$ denotes training-free methods.}
\vspace{0.2cm}
\label{tab:refinement_results}
\begin{tabular}{l|ccccc|c}
\toprule
\textbf{Method} & \textbf{AIME} & \textbf{AMC} & \textbf{MATH500} & \textbf{Minerva} & \textbf{OlympiadBench} & \textbf{Average} \\
\midrule
\multicolumn{7}{l}{\textbf{Qwen3-4B}} \\
\quad Base Model & 45.25 & 78.10 & 92.82 & 40.81 & 63.52 & 64.10 \\
\quad Reflexion$^\dagger$ & 38.47 & 73.17 & 91.48 & 40.87 & 60.89 & 60.98 \\
\quad Self-Refine$^\dagger$ & \underline{50.37} & 82.40 & 93.86 & 41.19 & 66.33 & 66.83 \\
\cmidrule(lr){1-7}
\quad GRPO & 48.91 & 81.86 & 93.78 & 42.90 & 69.67 & 67.42 \\
\quad DrGRPO & 46.98 & 82.66 & 94.46 & \textbf{44.84} & 71.75 & 68.14 \\
\quad DAPO & 49.86 & \textbf{87.31} & \underline{94.96} & 40.09 & \underline{72.81} & \underline{69.01} \\
\quad \textbf{\modelname{}} & \textbf{60.43} & \underline{85.54} & \textbf{95.70} & \underline{43.93} & \textbf{73.78} & \textbf{71.88} \\
\midrule
\multicolumn{7}{l}{\textbf{OLMo3-7B}} \\
\quad Base Model & 39.31 & 78.18 & 91.75 & 41.58 & 66.14 & 63.39 \\
\quad Reflexion$^\dagger$ & 37.38 & 72.18 & 91.29 & 41.48 & 62.36 & 60.94 \\
\quad Self-Refine$^\dagger$ & \underline{47.34} & 83.81 & 93.24 & 42.35 & 68.30 & \underline{67.01} \\
\cmidrule(lr){1-7}
\quad GRPO & 46.04 & 84.48 & 92.28 & 41.08 & 66.53 & 66.08 \\
\quad DrGRPO & 45.24 & 82.32 & \underline{93.54} & 42.75 & \underline{69.81} & 66.73 \\
\quad DAPO & 44.26 & \underline{84.55} & 93.33 & \underline{42.81} & 68.51 & 66.69 \\
\quad \textbf{\modelname{}} & \textbf{49.33} & \textbf{87.06} & \textbf{94.66} & \textbf{44.33} & \textbf{71.38} & \textbf{69.35} \\
\bottomrule
\end{tabular}
\end{table*}

\xhdr{Self-Refinement} We first evaluate the end-to-end reason-then-refine pipeline: for each problem, we sample $n{=}32$ base solutions,
refine each with greedy decoding (thus yielding $n{=}32$ refined solutions), and report pass@4 over the refined solutions. Table~\ref{tab:refinement_results} reports results on Qwen3-4B and Olmo3-7B. For Qwen3-4B, \modelname{} achieves the highest average of 71.88\%, outperforming DAPO by 2.9 percentage points and GRPO by 4.5 points. On AIME, \modelname{} reaches 60.43\%, a gain of over 11 percentage points above GRPO. For OLMo3-7B, the pattern is consistent: \modelname{} again achieves the highest average self-refinement score, outperforming all baselines on every benchmark. Meanwhile, \modelname{} also significantly outperforms training-free baselines\footnote{Even under our favorable implementation (details in Appendix \ref{app:baseline_impl}), Reflexion appears less suited to offline math refinement, because retrying from scratch after reflection can discard useful local corrections.}. The full self-refinement pass@$k$ curves are provided in Figure~\ref{fig:passatk_reflection} at Appendix~\ref{app:additional}.

\begin{figure}
    \centering
    \includegraphics[width=0.98\linewidth]{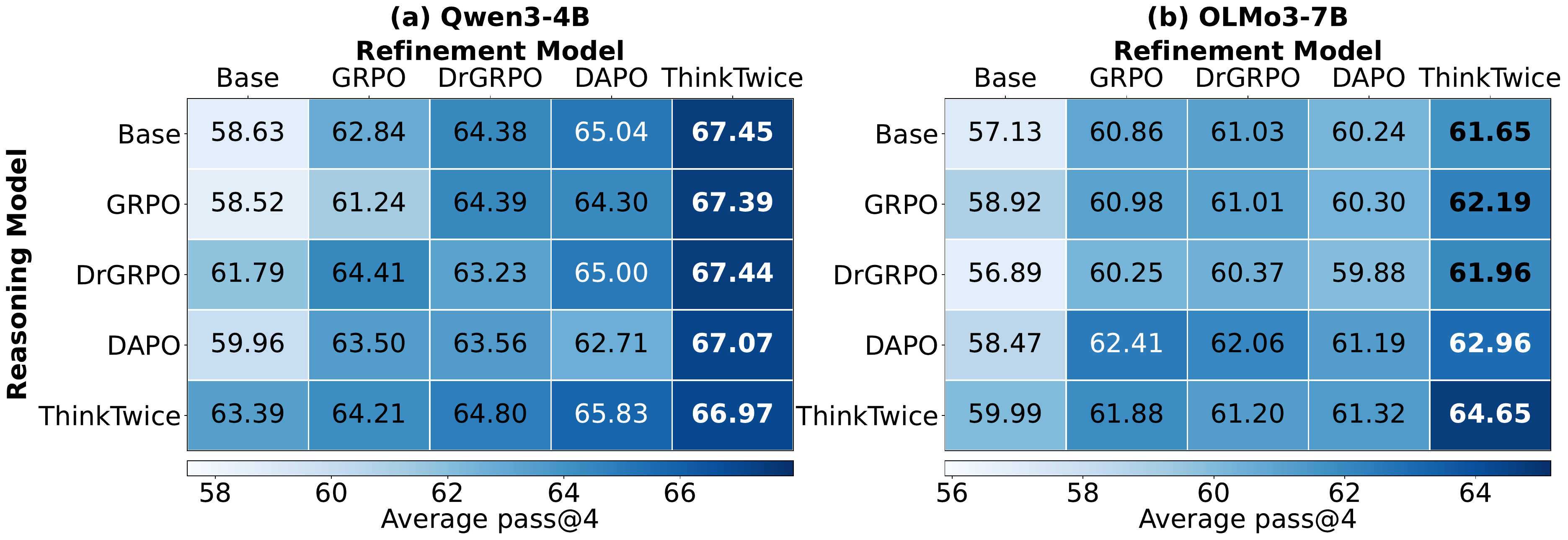}
    \caption{Cross-model refinement evaluation (average pass@4, $\uparrow$). Rows denote the backbone reasoning model; columns denote the refinement model.}
    \label{fig:cross_reflection}
\end{figure}

\xhdr{Cross-Model Refinement} To isolate refinement capability from base solution quality, we evaluate each model as a refinement model applied to every other model's base solutions. For each problem, we generate a single base solution with greedy decoding from the reasoning model, then sample 4 refinements from the refinement model and report the accuracy after refinement. This protocol fixes the input and attributes all variance to the model's refinement capabilities. Figure~\ref{fig:cross_reflection} reports the results averaged across all five benchmarks. \modelname{} as the refinement model (the rightmost column) achieves the highest score regardless of which model produced the base solution, demonstrating that its refinement capability generalizes beyond its own outputs. 

Taken together, these results show that \modelname{} improves both reasoning and self-refinement within a single training framework, and that these gains persist across two different types and sizes of LLMs.

\section{Discussion}

\subsection{Training Dynamics of \modelname{}}
\label{sec:dynamics}

We provide quantitative and qualitative analyses for understanding the learning dynamics of \modelname{}. Our central observation is that \modelname{} gives rise to an implicit \emph{rectify-then-fortify} dynamic: refinement predominantly corrects errors early in training, and gradually shifts toward preserving and polishing already-correct solutions as the base policy improves.

\begin{figure}[t]
    \centering
    \begin{subfigure}[t]{0.49\columnwidth}
        \centering
        \includegraphics[width=\linewidth]{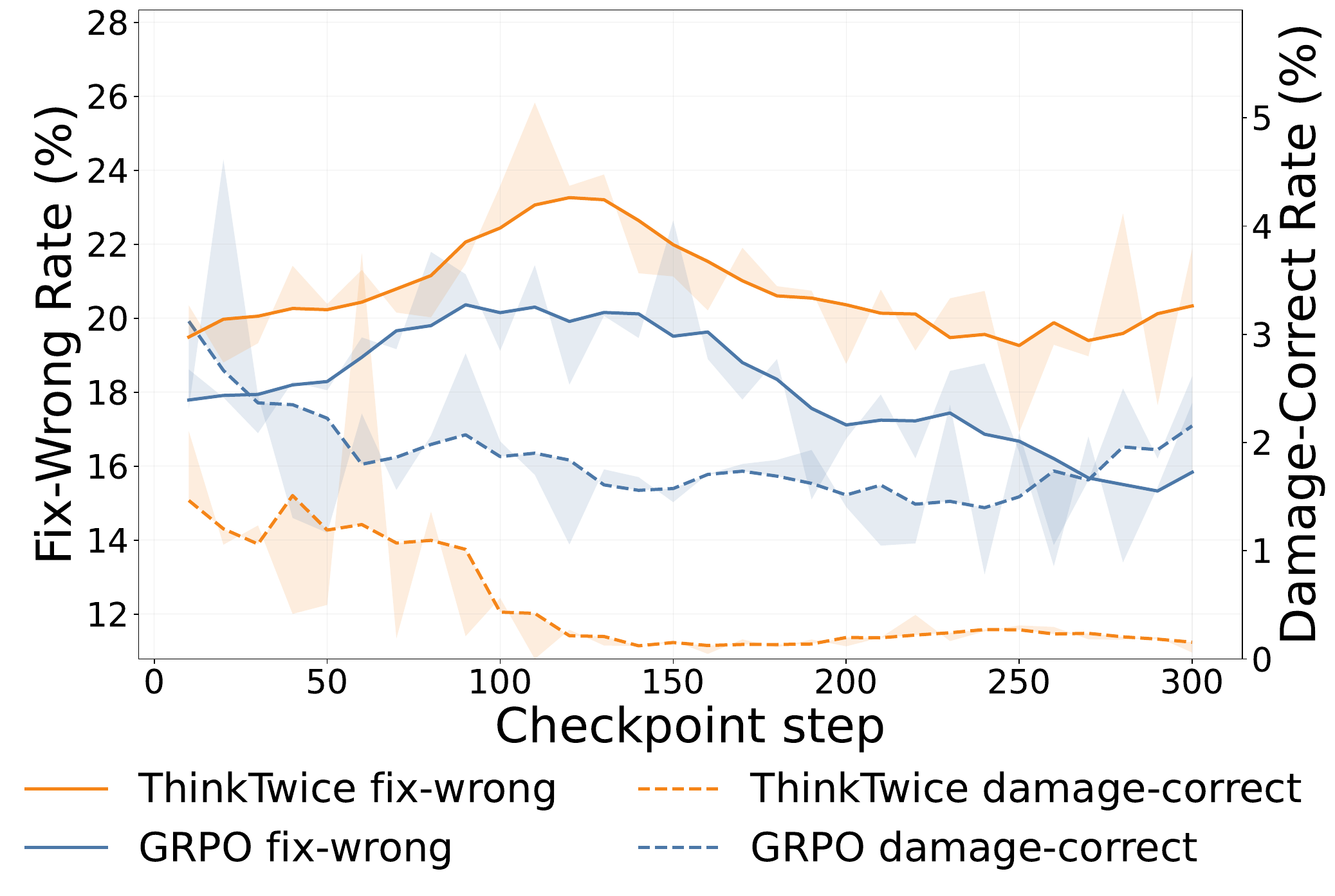}
        \phantomsubcaption\label{fig:transitions}
    \end{subfigure}\hfill
    \begin{subfigure}[t]{0.49\columnwidth}
        \centering
        \includegraphics[width=\linewidth]{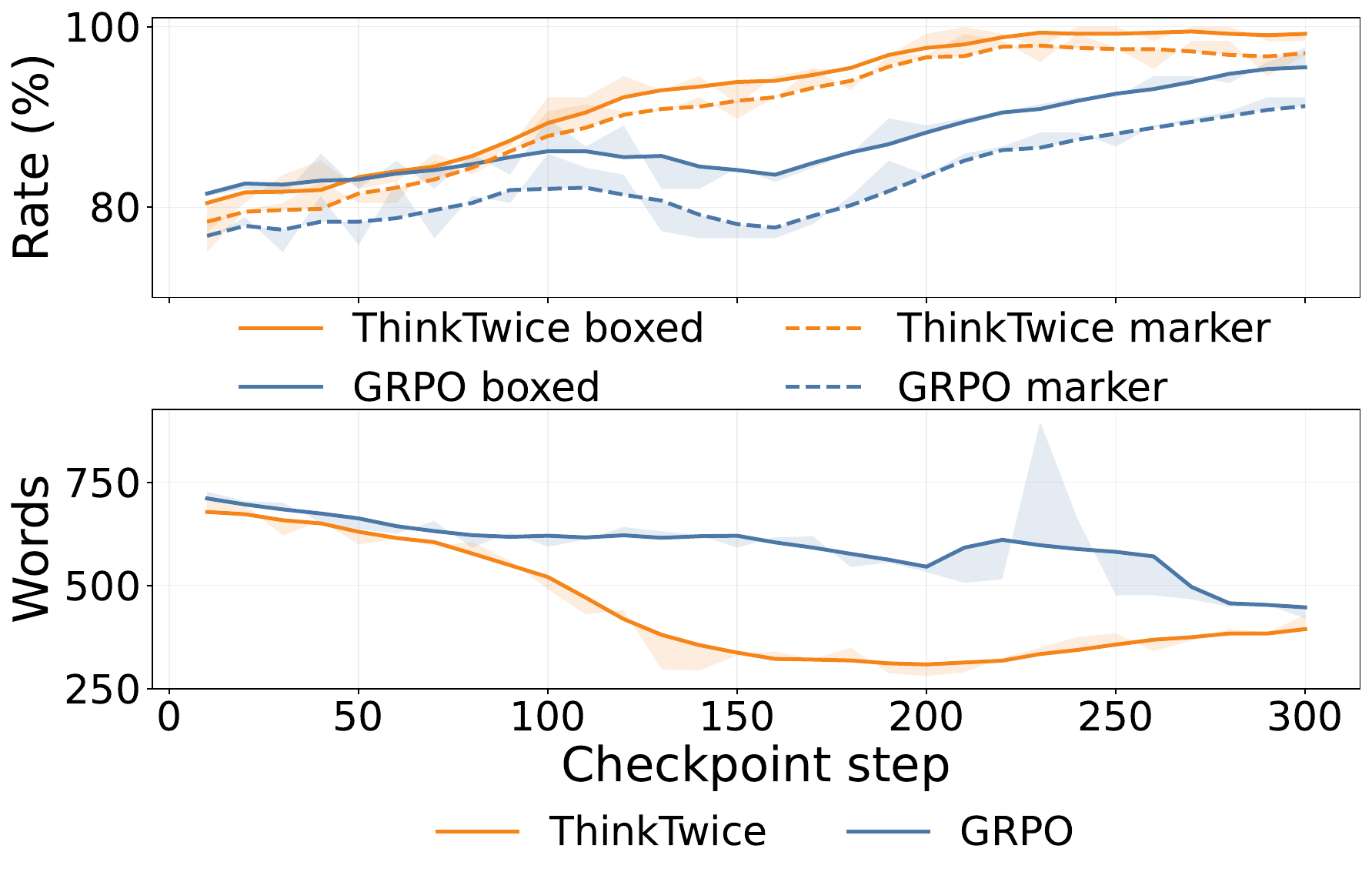}
        \phantomsubcaption\label{fig:polish_metrics}
    \end{subfigure}
    \caption{Training dynamics of refinement across checkpoints. The vertical dashed lines mark the best checkpoints. Left (a): transition metrics on the training set across checkpoints. Right (b): formatting and length metrics across checkpoints during training. Top: boxed-answer and final-answer marker rates; bottom: average response length for self-refinement on correct-only base solutions.}
    \label{fig:training_dynamics}
\end{figure}

\xhdr{Transition Metrics} To characterize this dynamic, we first define two conditional transition rates measured on the training set throughout the course of training. Let \emph{fix-wrong} (\%) denote the fraction of incorrect base solutions that become correct after refinement, and \emph{damage-correct} (\%) denote the fraction of correct base solutions that become incorrect after refinement. Figure~\ref{fig:training_dynamics}\subref{fig:transitions} plots both metrics across training checkpoints for \modelname{} and the GRPO baseline.

\xhdr{Early Training: Rectification via Refinement} In the early training
stages, the model can fail on mathematical problems due to typical failure modes, namely including incomplete derivations or unproductive search and solution paths. The refinement phase provides a second attempt on the same problem, allowing the model to accordingly recover from such failures where the base attempt narrowly fails. As shown in Figure~\ref{fig:training_dynamics}\subref{fig:transitions}, \modelname{} maintains a consistently higher \emph{fix-wrong} rate than the baseline throughout training, peaking around the half journey of the training. This means that for problems near the model's current capability boundary, the refinement phase can still produce correct solutions, yielding a more \emph{rectified} reward
signal than base reasoning alone would provide.

\xhdr{Late Training: Fortification via Refinement} As training progresses and the base policy becomes stronger and closer to model's capability boundary, fewer base solutions are incorrect, and the refinement phase naturally shifts from error correction to solution preservation. Figure~\ref{fig:training_dynamics}\subref{fig:transitions} shows that
\modelname{}'s \emph{damage-correct} rate drops near zero in the second half of training, while the baseline's rate is consistently more than 5 times higher than \modelname{}'s best checkpoint. Meanwhile, with \modelname{}, the model's outputs of correct solutions become significantly shorter over training (Figure~\ref{fig:training_dynamics}\subref{fig:polish_metrics}, Bottom), indicating that refinement is not merely acting as a prolonged generation window but rather helping remove exploratory clutter from already-correct solutions. Also, refined solutions exhibit better answer formatting than vanilla GRPO (Figure~\ref{fig:training_dynamics}\subref{fig:polish_metrics}, Top), with a higher rate of providing both boxed answers and \textit{Final Answer} markers despite no format reward being applied during training.

\xhdr{Qualitative Evidence} Finally, we provide detailed case studies in Appendix~\ref{app:case_studies} illustrating three recurring refinement behaviors that emerge during \modelname{} training: (i) \emph{route switching}, where refinement abandons a bad solution path and finds a better method; (ii) \emph{solution completion}, where a promising but unfinished base trace is completed by the refinement; and (iii) \emph{late-stage fortification}, where an already-correct base solution is shortened and cleaned up without altering the answer. These behaviors align with the quantitative transition from correction-dominant to preservation-dominant refinement observed above.

Taken together, the training dynamics suggest that \modelname{}'s joint optimization naturally yields a more rectified reward signal, as the refinement phase provides useful gradient
information from problems where the base attempt alone provides
limited learning signal.

\subsection{Training Cost of \modelname{}}
\label{sec:training_cost}

\begin{figure}[t]
    \centering
    \includegraphics[width=0.98\linewidth]{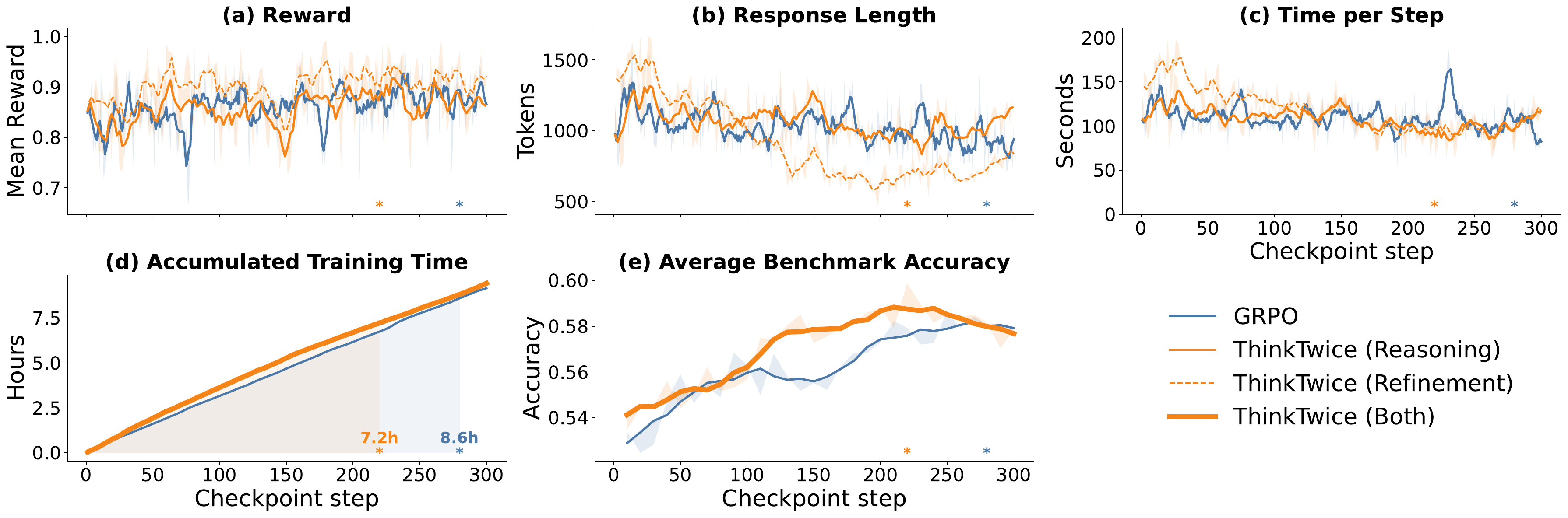}
    \caption{
    Training-time cost and dynamics of \modelname{} compared with GRPO. * denoted best checkpoint step of models.
    (a) Mean reward.
    (b) Response length.
    (c) Wall-clock time per update.
    (d) Accumulated training time.
    (e) Within-checkpoint macro average benchmark accuracy.
    Solid orange denotes \modelname{} base updates, solid blue denotes GRPO, and dashed
    orange denotes \modelname{} refinement updates when applicable.
    }
    \label{fig:training_cost_curves}
\end{figure}

A natural concern is whether the two-phase design of \modelname{}
introduces significant overhead or instabilities during training. We summarize core training-time signals from Qwen3-4B comparing \modelname{} with GRPO in Figure~\ref{fig:training_cost_curves}.  

Panel~(a) shows that the mean reward of \modelname{} remains comparably stable to GRPO throughout training, indicating that the refinement phase does not destabilize policy optimization. Panel~(b) shows that the base response lengths of \modelname{} stay in the same range as GRPO, while the refinement responses become progressively shorter over training, consistent with the shift toward concise and fortified generation observed in Section~\ref{sec:dynamics}.

Panel~(c) shows the wall-clock time per update\footnote{All timing results are measured on 2$\times$H100 80GB GPUs. Hyperparameter details are in Appendix~\ref{app:imp_details}.}. Refinement phase updates are more expensive than standard GRPO updates, especially early in training, because the refinement prompt prepends the model's earlier solution and therefore contains substantially longer context. However, as training progresses, refined responses become
progressively shorter (as shown in Panel~(b)), narrowing the
training cost gap. Thus, Panel~(d) shows that the accumulated wall-clock time of \modelname{} remains close to GRPO: at the same step count up to step 300, \modelname{} is only 3\% slower in total wall-clock time (9.42 hours vs 9.15 hours). Moreover, \modelname{} reaches its best checkpoint in 16\% less wall-clock time than GRPO (7.2h vs 8.6h), as it converges in fewer total steps (220 vs 280). Finally, Panel~(e) shows that \modelname{} improves faster in the early and middle stages of training and maintains higher benchmark
accuracy than GRPO for most of the trajectory.

\section{Conclusion}

We introduced \modelname{}, a two-phase RLVR framework that jointly optimizes reasoning and self-refinement using the same
binary correctness reward in both phases, without correctness signals, critique annotations, or external verifiers. Across five mathematical reasoning benchmarks and two model families, \modelname{} consistently outperforms competitive online policy optimization baselines on both direct reasoning and self-refinement. Our quantitative and qualitative analyses of the training dynamics of \modelname{} reveal an implicit rectify-then-fortify curriculum that yields a more rectified reward signal, while adding only minimal training overhead compared to GRPO which we build upon.

Several directions naturally extend this work. First, while we
evaluate on mathematical reasoning with verifiable answers,
\modelname{}'s design is domain-agnostic and could be applied to
other tasks with outcome-based rewards, such as code generation.
Second, our framework natively supports arbitrary numbers of
refinement turns within the multi-turn conversation format;
exploring multi-step iterative refinement is a direct extension.

\section*{Ethics Statement}
This work focuses on improving mathematical reasoning capabilities
of large language models through reinforcement learning with
verifiable rewards. Our method uses publicly available datasets and
models, and does not involve human subjects, private data, or
dual-use concerns beyond those inherent to general-purpose language
model research.

\section*{Reproducibility Statement}
We provide comprehensive implementation details to facilitate
reproduction of our results. Appendix~\ref{app:imp_details} includes
the full training workflow with code-level illustrations, the exact
refinement instruction, hyperparameter configurations
(Table~\ref{tab:hparams}), and dataset sources with links.
Appendix~\ref{app:baseline_impl} details the implementation of all
training-free baselines. An anonymous codebase for \modelname{} is linked in the abstract. All experiments use publicly available base models and datasets.

\clearpage

\bibliography{colm2026_conference}

@article{hu2025open,
  title={Open-reasoner-zero: An open source approach to scaling up reinforcement learning on the base model},
  author={Hu, Jingcheng and Zhang, Yinmin and Han, Qi and Jiang, Daxin and Zhang, Xiangyu and Shum, Heung-Yeung},
  journal={arXiv preprint arXiv:2503.24290},
  year={2025}
}

@article{zeng2025simplerl,
  title={Simplerl-zoo: Investigating and taming zero reinforcement learning for open base models in the wild},
  author={Zeng, Weihao and Huang, Yuzhen and Liu, Qian and Liu, Wei and He, Keqing and Ma, Zejun and He, Junxian},
  journal={arXiv preprint arXiv:2503.18892},
  year={2025}
}

@article{he2025skywork,
  title={Skywork open reasoner 1 technical report},
  author={He, Jujie and Liu, Jiacai and Liu, Chris Yuhao and Yan, Rui and Wang, Chaojie and Cheng, Peng and Zhang, Xiaoyu and Zhang, Fuxiang and Xu, Jiacheng and Shen, Wei and others},
  journal={arXiv preprint arXiv:2505.22312},
  year={2025}
}

@article{fu2025areal,
  title={Areal: A large-scale asynchronous reinforcement learning system for language reasoning},
  author={Fu, Wei and Gao, Jiaxuan and Shen, Xujie and Zhu, Chen and Mei, Zhiyu and He, Chuyi and Xu, Shusheng and Wei, Guo and Mei, Jun and Wang, Jiashu and others},
  journal={arXiv preprint arXiv:2505.24298},
  year={2025}
}

@article{zheng2025group,
  title={Group sequence policy optimization},
  author={Zheng, Chujie and Liu, Shixuan and Li, Mingze and Chen, Xiong-Hui and Yu, Bowen and Gao, Chang and Dang, Kai and Liu, Yuqiong and Men, Rui and Yang, An and others},
  journal={arXiv preprint arXiv:2507.18071},
  year={2025}
}

@article{zhao2025geometric,
  title={Geometric-mean policy optimization},
  author={Zhao, Yuzhong and Liu, Yue and Liu, Junpeng and Chen, Jingye and Wu, Xun and Hao, Yaru and Lv, Tengchao and Huang, Shaohan and Cui, Lei and Ye, Qixiang and others},
  journal={arXiv preprint arXiv:2507.20673},
  year={2025}
}

@article{chu2025gpg,
  title={Gpg: A simple and strong reinforcement learning baseline for model reasoning},
  author={Chu, Xiangxiang and Huang, Hailang and Zhang, Xiao and Wei, Fei and Wang, Yong},
  journal={arXiv preprint arXiv:2504.02546},
  year={2025}
}

@article{zeng2025shrinking,
  title={Shrinking the Variance: Shrinkage Baselines for Reinforcement Learning with Verifiable Rewards},
  author={Zeng, Guanning and Zhou, Zhaoyi and Arora, Daman and Zanette, Andrea},
  journal={arXiv preprint arXiv:2511.03710},
  year={2025}
}

@article{liu2025understanding,
  title={Understanding r1-zero-like training: A critical perspective},
  author={Liu, Zichen and Chen, Changyu and Li, Wenjun and Qi, Penghui and Pang, Tianyu and Du, Chao and Lee, Wee Sun and Lin, Min},
  journal={arXiv preprint arXiv:2503.20783},
  year={2025}
}

@article{guo2025deepseek,
  title={Deepseek-r1: Incentivizing reasoning capability in llms via reinforcement learning},
  author={Guo, Daya and Yang, Dejian and Zhang, Haowei and Song, Junxiao and Wang, Peiyi and Zhu, Qihao and Xu, Runxin and Zhang, Ruoyu and Ma, Shirong and Bi, Xiao and others},
  journal={arXiv preprint arXiv:2501.12948},
  year={2025}
}

@article{shao2024deepseekmath,
  title={Deepseekmath: Pushing the limits of mathematical reasoning in open language models},
  author={Shao, Zhihong and Wang, Peiyi and Zhu, Qihao and Xu, Runxin and Song, Junxiao and Bi, Xiao and Zhang, Haowei and Zhang, Mingchuan and Li, YK and Wu, Yang and others},
  journal={arXiv preprint arXiv:2402.03300},
  year={2024}
}

@article{madaan2023self,
  title={Self-refine: Iterative refinement with self-feedback},
  author={Madaan, Aman and Tandon, Niket and Gupta, Prakhar and Hallinan, Skyler and Gao, Luyu and Wiegreffe, Sarah and Alon, Uri and Dziri, Nouha and Prabhumoye, Shrimai and Yang, Yiming and others},
  journal={Advances in neural information processing systems},
  volume={36},
  pages={46534--46594},
  year={2023}
}

@article{shinn2023reflexion,
  title={Reflexion: Language agents with verbal reinforcement learning},
  author={Shinn, Noah and Cassano, Federico and Gopinath, Ashwin and Narasimhan, Karthik and Yao, Shunyu},
  journal={Advances in neural information processing systems},
  volume={36},
  pages={8634--8652},
  year={2023}
}

@article{gou2023critic,
  title={Critic: Large language models can self-correct with tool-interactive critiquing},
  author={Gou, Zhibin and Shao, Zhihong and Gong, Yeyun and Shen, Yelong and Yang, Yujiu and Duan, Nan and Chen, Weizhu},
  journal={arXiv preprint arXiv:2305.11738},
  year={2023}
}

@misc{Wang2022SelfConsistency,
  title         = {Self-Consistency Improves Chain of Thought Reasoning in Language Models},
  author        = {Wang, Xuezhi and Wei, Jason and Schuurmans, Dale and Le, Quoc and Chi, Ed and Narang, Sharan and Chowdhery, Aakanksha and Zhou, Denny},
  year          = {2022},
  eprint        = {2203.11171},
  archivePrefix = {arXiv},
  primaryClass  = {cs.CL},
  doi           = {10.48550/arXiv.2203.11171},
  url           = {https://arxiv.org/abs/2203.11171},
  note          = {ICLR 2023}
}

@article{huang2023large,
  title={Large language models cannot self-correct reasoning yet},
  author={Huang, Jie and Chen, Xinyun and Mishra, Swaroop and Zheng, Huaixiu Steven and Yu, Adams Wei and Song, Xinying and Zhou, Denny},
  journal={arXiv preprint arXiv:2310.01798},
  year={2023}
}

@inproceedings{weng2023large,
  title={Large language models are better reasoners with self-verification},
  author={Weng, Yixuan and Zhu, Minjun and Xia, Fei and Li, Bin and He, Shizhu and Liu, Shengping and Sun, Bin and Liu, Kang and Zhao, Jun},
  booktitle={Findings of the Association for Computational Linguistics: EMNLP 2023},
  pages={2550--2575},
  year={2023}
}

@article{welleck2022generating,
  title={Generating sequences by learning to self-correct},
  author={Welleck, Sean and Lu, Ximing and West, Peter and Brahman, Faeze and Shen, Tianxiao and Khashabi, Daniel and Choi, Yejin},
  journal={arXiv preprint arXiv:2211.00053},
  year={2022}
}

@article{qu2024recursive,
  title={Recursive introspection: Teaching language model agents how to self-improve},
  author={Qu, Yuxiao and Zhang, Tianjun and Garg, Naman and Kumar, Aviral},
  journal={Advances in Neural Information Processing Systems},
  volume={37},
  pages={55249--55285},
  year={2024}
}

@inproceedings{zhang2024small,
  title={Small language models need strong verifiers to self-correct reasoning},
  author={Zhang, Yunxiang and Khalifa, Muhammad and Logeswaran, Lajanugen and Kim, Jaekyeom and Lee, Moontae and Lee, Honglak and Wang, Lu},
  booktitle={Findings of the Association for Computational Linguistics: ACL 2024},
  pages={15637--15653},
  year={2024}
}

@article{xiong2025self,
  title={Self-rewarding correction for mathematical reasoning},
  author={Xiong, Wei and Zhang, Hanning and Ye, Chenlu and Chen, Lichang and Jiang, Nan and Zhang, Tong},
  journal={arXiv preprint arXiv:2502.19613},
  year={2025}
}

@article{zhao2025boosting,
  title={Boosting llm reasoning via spontaneous self-correction},
  author={Zhao, Xutong and Xu, Tengyu and Wang, Xuewei and Chen, Zhengxing and Jin, Di and Tan, Liang and Yu, Zishun and Zhao, Zhuokai and He, Yun and Wang, Sinong and others},
  journal={arXiv preprint arXiv:2506.06923},
  year={2025}
}

@article{cobbe2021training,
  title={Training verifiers to solve math word problems},
  author={Cobbe, Karl and Kosaraju, Vineet and Bavarian, Mohammad and Chen, Mark and Jun, Heewoo and Kaiser, Lukasz and Plappert, Matthias and Tworek, Jerry and Hilton, Jacob and Nakano, Reiichiro and others},
  journal={arXiv preprint arXiv:2110.14168},
  year={2021}
}

@article{uesato2022solving,
  title={Solving math word problems with process-and outcome-based feedback},
  author={Uesato, Jonathan and Kushman, Nate and Kumar, Ramana and Song, Francis and Siegel, Noah and Wang, Lisa and Creswell, Antonia and Irving, Geoffrey and Higgins, Irina},
  journal={arXiv preprint arXiv:2211.14275},
  year={2022}
}

@inproceedings{lightman2023let,
  title={Let's verify step by step},
  author={Lightman, Hunter and Kosaraju, Vineet and Burda, Yuri and Edwards, Harrison and Baker, Bowen and Lee, Teddy and Leike, Jan and Schulman, John and Sutskever, Ilya and Cobbe, Karl},
  booktitle={The twelfth international conference on learning representations},
  year={2023}
}

@inproceedings{wang2024math,
  title={Math-shepherd: Verify and reinforce llms step-by-step without human annotations},
  author={Wang, Peiyi and Li, Lei and Shao, Zhihong and Xu, Runxin and Dai, Damai and Li, Yifei and Chen, Deli and Wu, Yu and Sui, Zhifang},
  booktitle={Proceedings of the 62nd Annual Meeting of the Association for Computational Linguistics (Volume 1: Long Papers)},
  pages={9426--9439},
  year={2024}
}

@article{yuan2024free,
  title={Free process rewards without process labels},
  author={Yuan, Lifan and Li, Wendi and Chen, Huayu and Cui, Ganqu and Ding, Ning and Zhang, Kaiyan and Zhou, Bowen and Liu, Zhiyuan and Peng, Hao},
  journal={arXiv preprint arXiv:2412.01981},
  year={2024}
}

@article{cui2025process,
  title={Process reinforcement through implicit rewards},
  author={Cui, Ganqu and Yuan, Lifan and Wang, Zefan and Wang, Hanbin and Zhang, Yuchen and Chen, Jiacheng and Li, Wendi and He, Bingxiang and Fan, Yuchen and Yu, Tianyu and others},
  journal={arXiv preprint arXiv:2502.01456},
  year={2025}
}

@article{xi2024enhancing,
  title={Enhancing LLM reasoning via critique models with test-time and training-time supervision},
  author={Xi, Zhiheng and Yang, Dingwen and Huang, Jixuan and Tang, Jiafu and Li, Guanyu and Ding, Yiwen and He, Wei and Hong, Boyang and Do, Shihan and Zhan, Wenyu and others},
  journal={arXiv preprint arXiv:2411.16579},
  year={2024}
}

@inproceedings{yu2025training,
  title={Training language model to critique for better refinement},
  author={Yu, Tianshu and Xiang, Chao and Yang, Mingchuan and Ke, Pei and Wen, Bosi and Wang, Cunxiang and Cheng, Jiale and Zhang, Li and Mu, Xinyu and Sun, Chuxiong and others},
  booktitle={Findings of the Association for Computational Linguistics: ACL 2025},
  pages={26760--26804},
  year={2025}
}

@article{kumar2024training,
  title={Training language models to self-correct via reinforcement learning},
  author={Kumar, Aviral and Zhuang, Vincent and Agarwal, Rishabh and Su, Yi and Co-Reyes, John D and Singh, Avi and Baumli, Kate and Iqbal, Shariq and Bishop, Colton and Roelofs, Rebecca and others},
  journal={arXiv preprint arXiv:2409.12917},
  year={2024}
}

@inproceedings{ma2025s2r,
  title={S2r: Teaching llms to self-verify and self-correct via reinforcement learning},
  author={Ma, Ruotian and Wang, Peisong and Liu, Cheng and Liu, Xingyan and Chen, Jiaqi and Zhang, Bang and Zhou, Xin and Du, Nan and Li, Jia},
  booktitle={Proceedings of the 63rd Annual Meeting of the Association for Computational Linguistics (Volume 1: Long Papers)},
  pages={22632--22654},
  year={2025}
}

@inproceedings{he2025rise,
  title={Rise: reasoning enhancement via iterative self-exploration in multi-hop question answering},
  author={He, Bolei and He, Xinran and Chen, Mengke and Xue, Xianwei and Zhu, Ying and Ling, Zhen-Hua},
  booktitle={Findings of the Association for Computational Linguistics: ACL 2025},
  pages={14925--14948},
  year={2025}
}

@article{jiang2025pag,
  title={Pag: Multi-turn reinforced llm self-correction with policy as generative verifier},
  author={Jiang, Yuhua and Xiong, Yuwen and Yuan, Yufeng and Xin, Chao and Xu, Wenyuan and Yue, Yu and Zhao, Qianchuan and Yan, Lin},
  journal={arXiv preprint arXiv:2506.10406},
  year={2025}
}

@article{zhang2025incentivizing,
  title={Incentivizing LLMs to Self-Verify Their Answers},
  author={Zhang, Fuxiang and Xu, Jiacheng and Wang, Chaojie and Cui, Ce and Liu, Yang and An, Bo},
  journal={arXiv preprint arXiv:2506.01369},
  year={2025}
}

@article{zhang2025critique,
  title={Critique-grpo: Advancing llm reasoning with natural language and numerical feedback},
  author={Zhang, Xiaoying and Zhang, Yipeng and Sun, Hao and Feng, Kaituo and Lu, Chaochao and Yang, Chao and Meng, Helen},
  journal={arXiv preprint arXiv:2506.03106},
  year={2025}
}

@misc{Yu2025DAPO,
  title         = {{DAPO}: An Open-Source {LLM} Reinforcement Learning System at Scale},
  author        = {Yu, Qiying and Zhang, Zheng and Zhu, Ruofei and Yuan, Yufeng and Zuo, Xiaochen and Yue, Yu and Dai, Weinan and Fan, Tiantian and Liu, Gaohong and others},
  year          = {2025},
  eprint        = {2503.14476},
  archivePrefix = {arXiv},
  primaryClass  = {cs.LG},
  doi           = {10.48550/arXiv.2503.14476},
  url           = {https://arxiv.org/abs/2503.14476}
}

@misc{Zhang2024ReSTMCTS,
  title         = {{ReST}-{MCTS}*: {LLM} Self-Training via Process Reward Guided Tree Search},
  author        = {Zhang, Dan and Zhoubian, Sining and Hu, Ziniu and Yue, Yisong and Dong, Yuxiao and Tang, Jie},
  year          = {2024},
  eprint        = {2406.03816},
  archivePrefix = {arXiv},
  primaryClass  = {cs.CL},
  doi           = {10.48550/arXiv.2406.03816},
  url           = {https://arxiv.org/abs/2406.03816},
  note          = {NeurIPS 2024}
}

@article{hendrycks2021measuring,
  title={Measuring mathematical problem solving with the math dataset},
  author={Hendrycks, Dan and Burns, Collin and Kadavath, Saurav and Arora, Akul and Basart, Steven and Tang, Eric and Song, Dawn and Steinhardt, Jacob},
  journal={arXiv preprint arXiv:2103.03874},
  year={2021}
}

@article{lewkowycz2022solving,
  title={Solving quantitative reasoning problems with language models},
  author={Lewkowycz, Aitor and Andreassen, Anders and Dohan, David and Dyer, Ethan and Michalewski, Henryk and Ramasesh, Vinay and Slone, Ambrose and Anil, Cem and Schlag, Imanol and Gutman-Solo, Theo and others},
  journal={Advances in neural information processing systems},
  volume={35},
  pages={3843--3857},
  year={2022}
}

@inproceedings{he2024olympiadbench,
  title={Olympiadbench: A challenging benchmark for promoting agi with olympiad-level bilingual multimodal scientific problems},
  author={He, Chaoqun and Luo, Renjie and Bai, Yuzhuo and Hu, Shengding and Thai, Zhen and Shen, Junhao and Hu, Jinyi and Han, Xu and Huang, Yujie and Zhang, Yuxiang and others},
  booktitle={Proceedings of the 62nd Annual Meeting of the Association for Computational Linguistics (Volume 1: Long Papers)},
  pages={3828--3850},
  year={2024}
}

@article{yang2025qwen3,
  title={Qwen3 technical report},
  author={Yang, An and Li, Anfeng and Yang, Baosong and Zhang, Beichen and Hui, Binyuan and Zheng, Bo and Yu, Bowen and Gao, Chang and Huang, Chengen and Lv, Chenxu and others},
  journal={arXiv preprint arXiv:2505.09388},
  year={2025}
}

@article{olmo2025olmo,
  title={Olmo 3},
  author={Olmo, Team and Ettinger, Allyson and Bertsch, Amanda and Kuehl, Bailey and Graham, David and Heineman, David and Groeneveld, Dirk and Brahman, Faeze and Timbers, Finbarr and Ivison, Hamish and others},
  journal={arXiv preprint arXiv:2512.13961},
  year={2025}
}

@article{sheng2024hybridflow,
  title   = {HybridFlow: A Flexible and Efficient RLHF Framework},
  author  = {Guangming Sheng and Chi Zhang and Zilingfeng Ye and Xibin Wu and Wang Zhang and Ru Zhang and Yanghua Peng and Haibin Lin and Chuan Wu},
  year    = {2024},
  journal = {arXiv preprint arXiv: 2409.19256}
}

@article{chen2021evaluating,
  title={Evaluating large language models trained on code},
  author={Chen, Mark and Tworek, Jerry and Jun, Heewoo and Yuan, Qiming and Pinto, Henrique Ponde De Oliveira and Kaplan, Jared and Edwards, Harri and Burda, Yuri and Joseph, Nicholas and Brockman, Greg and others},
  journal={arXiv preprint arXiv:2107.03374},
  year={2021}
}

@book{polya1945solve,
  title={How to solve it: A new aspect of mathematical method},
  author={Polya, George},
  year={1945},
  publisher={Princeton university press}
}

@article{burns2023weak,
  title={Weak-to-strong generalization: Eliciting strong capabilities with weak supervision},
  author={Burns, Collin and Izmailov, Pavel and Kirchner, Jan Hendrik and Baker, Bowen and Gao, Leo and Aschenbrenner, Leopold and Chen, Yining and Ecoffet, Adrien and Joglekar, Manas and Leike, Jan and others},
  journal={arXiv preprint arXiv:2312.09390},
  year={2023}
}

@article{tian2025think,
  title={Think twice: Enhancing llm reasoning by scaling multi-round test-time thinking},
  author={Tian, Xiaoyu and Zhao, Sitong and Wang, Haotian and Chen, Shuaiting and Ji, Yunjie and Peng, Yiping and Zhao, Han and Li, Xiangang},
  journal={arXiv preprint arXiv:2503.19855},
  year={2025}
}

@article{jiao2025think,
  title={Think Twice: Branch-and-Rethink Reasoning Reward Model},
  author={Jiao, Yizhu and Zeng, Jiaqi and Vialard, Julien Veron and Kuchaiev, Oleksii and Han, Jiawei and Delalleau, Olivier},
  journal={arXiv preprint arXiv:2510.23596},
  year={2025}
}

@article{qian2023think,
  title={Think twice: A human-like two-stage conversational agent for emotional response generation},
  author={Qian, Yushan and Wang, Bo and Ma, Shangzhao and Bin, Wu and Zhang, Shuo and Zhao, Dongming and Huang, Kun and Hou, Yuexian},
  journal={arXiv preprint arXiv:2301.04907},
  year={2023}
}

@inproceedings{wilf2024think,
  title={Think twice: Perspective-taking improves large language models’ theory-of-mind capabilities},
  author={Wilf, Alex and Lee, Sihyun and Liang, Paul Pu and Morency, Louis-Philippe},
  booktitle={Proceedings of the 62nd Annual Meeting of the Association for Computational Linguistics (Volume 1: Long Papers)},
  pages={8292--8308},
  year={2024}
}

@inproceedings{mikula2024think,
  title={Think twice: Measuring the efficiency of eliminating prediction shortcuts of question answering models},
  author={Mikula, Luk{\'a}{\v{s}} and {\v{S}}tef{\'a}nik, Michal and Petrovi{\v{c}}, Marek and Sojka, Petr},
  booktitle={Proceedings of the 18th Conference of the European Chapter of the Association for Computational Linguistics (Volume 1: Long Papers)},
  pages={2179--2193},
  year={2024}
}

@inproceedings{li2024think,
  title={Think twice before trusting: Self-detection for large language models through comprehensive answer reflection},
  author={Li, Moxin and Wang, Wenjie and Feng, Fuli and Zhu, Fengbin and Wang, Qifan and Chua, Tat-Seng},
  booktitle={Findings of the Association for Computational Linguistics: EMNLP 2024},
  pages={11858--11875},
  year={2024}
}

@article{phan2025think,
  title={Think Twice, Generate Once: Safeguarding by Progressive Self-Reflection},
  author={Phan, Hoang and Li, Victor and Lei, Qi},
  journal={arXiv preprint arXiv:2510.01270},
  year={2025}
}
\bibliographystyle{colm2026_conference}

\appendix
\appendix
\onecolumn

\section{Additional Related Work Comparison}
\label{app:related-work-table-sec}

\begin{table*}[h]
\centering
\caption{A compact comparison of \modelname{} with related work. We compare whether methods use reinforcement learning, require supervision beyond final-answer correctness, and rely on an additional verifier, teacher, or explicit critique channel. Entries summarize the dominant tendency of each family.}
\label{tab:related_work_comparison}
\scriptsize
\renewcommand{\arraystretch}{1.08}
\setlength{\tabcolsep}{3pt}
\begin{tabularx}{\textwidth}{
>{\raggedright\arraybackslash}p{1.75cm}
>{\raggedright\arraybackslash}p{3.65cm}
>{\centering\arraybackslash}p{0.65cm}
>{\raggedright\arraybackslash}p{2.15cm}
>{\raggedright\arraybackslash}p{1.55cm}
>{\raggedright\arraybackslash}X}
\toprule
\textbf{Family}
& \textbf{Representative methods}
& \textbf{RL?}
& \textbf{Extra supervision?}
& \textbf{Extra verifier / teacher / critique}
& \textbf{Core characteristic} \\
\midrule

One-pass RLVR
& GRPO, DeepSeek-R1, DAPO, Dr.~GRPO
& Yes
& No
& No
& Trains first-pass reasoning only; no learned revision policy. \\

Training-free refinement
& Self-Refine, Reflexion, self-verification prompting, CRITIC, self-consistency
& No
& No during training
& No, or optional tools
& Uses extra test-time feedback or sampling, but refinement is not learned. \\

Training-based refinement
& Cobbe verifier, Let’s Verify, Math-Shepherd, Free Process Rewards, PRIME, AutoMathCritique, CFT, Recursive Introspection, Self-rewarding Correction, SPOC
& Mixed
& Often yes
& Often yes or partial
& Usually relies on process labels, critique data, synthetic correction traces, or stronger verifiers. \\

RL-based refinement
& SCoRe, S$^2$R, RISE, PAG, Self-Verify, Critique-GRPO
& Yes
& Usually yes
& Usually explicit
& Closest prior line, but typically adds explicit verification or critique objectives. \\

\midrule
\textbf{\modelname{}}
& Two-phase RLVR with a shared policy
& \textbf{Yes}
& \textbf{No}
& \textbf{No}
& \textbf{Same correctness reward in solve and revise phases; revision conditions only on the model's prior answer plus a generic review instruction.} \\

\bottomrule
\end{tabularx}
\end{table*}

\xhdr{Position of \modelname{} in the literature} Table~\ref{tab:related_work_comparison} provides a broader comparison between \modelname{} and prior approaches discussed in Section~\ref{sec:related_work}. We organize the literature into four families: one-pass RLVR methods, training-free self-refinement methods, training-based self-refinement methods, and RL-based self-refinement methods. The comparison highlights that \modelname{} is closest to the RL-based self-refinement line, but differs in using a shared policy and the same sparse final-answer correctness reward in both the solve and revise phases, without process labels, critique annotations, correctness hints, or an explicit verifier indicating whether the initial answer is correct.

\xhdr{Other second-pass methods}
Prior work has explored second-pass reasoning, reflection, and related reconsideration strategies in a range of settings, including emotional dialogue generation~\citep{qian2023think}, Theory-of-Mind prompting via \emph{SimToM}~\citep{wilf2024think}, shortcut analysis in question answering~\citep{mikula2024think}, LLM self-detection via \emph{T3}~\citep{li2024think}, \emph{Multi-round Thinking} for inference-time scaling~\citep{tian2025think}, \emph{Progressive Self-Reflection} for safety~\citep{phan2025think}, and \emph{branch-and-rethink} for reward modeling~\citep{jiao2025think}. Although these works share the high-level intuition that revisiting an initial response can help, they differ substantially in task, objective, and supervision. In contrast, our \modelname{} jointly trains a single reasoning policy with RL to perform both first-pass reasoning and second-pass refinement using the same sparse final-answer correctness reward in both phases, without critique annotations, external verifiers, or a separate reward-model objective.

\section{Reproducibility}

\subsection{\modelname{} Implementation Details}
\label{app:imp_details}

\subsubsection{Datasets}

We use the MATH dataset~\footnote{\url{https://huggingface.co/datasets/hendrycks/math}} for training, which consists of 7,500 problems spanning difficulty levels 1--5 across seven subjects. For evaluation, we use a test set of 1,526 problems from five widely used mathematical reasoning benchmarks:

\begin{itemize}
    \item \textbf{MATH500}~\footnote{\url{https://huggingface.co/datasets/HuggingFaceH4/MATH-500}}: 500 problems from the MATH test set.
    \item \textbf{OlympiadBench}~\footnote{\url{https://huggingface.co/datasets/afraamn/olympiadbench_math_textonly}}: 581 problems from olympiad-level competitions.
    \item \textbf{Minerva Math}~\footnote{\url{https://huggingface.co/datasets/math-ai/minervamath}}: 272 problems from high school math competitions.
    \item \textbf{AIME}: 90 problems from AIME 2022--2024~\footnote{\url{https://huggingface.co/datasets/AI-MO/aimo-validation-aime}}.
    \item \textbf{AMC}: 83 problems from the AMC 10/12 competitions~\footnote{\url{https://huggingface.co/datasets/AI-MO/aimo-validation-amc}}.
\end{itemize}

Each problem is formatted with a task instruction requiring the model to output its final answer in boxed format (\texttt{\textbackslash boxed\{\}}). We evaluate accuracy using exact matching on the boxed content with Huggingface Math-Verify.

\subsubsection{A Detailed Illustration of \modelname{} Training Workflow}

We provide a concrete implementation-style walkthrough of \modelname{} training, bridging the Algorithm~\ref{alg:OurMethod} pseudocode with the actual code execution flow. The following uses Python-like notation containing the core training logic from \texttt{ray\_trainer.py}.

\xhdr{The two-phase training} The implementation follows the two-phase structure in Algorithm~\ref{alg:OurMethod}. When we activate \modelname{} training, the main training loop alternates between reasoning and refinement steps within each iteration:

\begin{lstlisting}[language=Python, basicstyle=\ttfamily\small, frame=single, breaklines=true, keywordstyle=\color{blue}\bfseries, stringstyle=\color{green!60!black}, commentstyle=\color{gray}\itshape, showstringspaces=false]
for epoch in range(total_epochs):
    for batch in dataloader:
        # Phase 1: Reasoning (with refinement candidate preparation)
        pending_refinement = run_step(batch, allow_refinement=True)
        
        # Phase 2: Refinement (if candidates selected)
        if pending_refinement:
            run_step(pending_refinement, allow_refinement=False)
\end{lstlisting}

When \texttt{allow\_refinement=True}, \texttt{run\_step} internally performs candidate selection after reward computation and policy update:

\begin{lstlisting}[language=Python, basicstyle=\ttfamily\small, frame=single, breaklines=true, keywordstyle=\color{blue}\bfseries, stringstyle=\color{green!60!black}, commentstyle=\color{gray}\itshape, showstringspaces=false]
num_select = len(batch) // repeat_times  # One per prompt
refinement_indices = select_refinement_indices(
    uids, responses, is_correct, num_select,
    selection_mode='random'  # for random selection from base solutions
)
selected_base = base_batch.select_idxs(refinement_indices)
selected_gen = gen_batch_output.select_idxs(refinement_indices)
# yielding (selected_base, selected_gen) for refinement prompt
\end{lstlisting}

The refinement prompts are then constructed from \texttt{selected\_gen} (containing the generated responses) and \texttt{selected\_base} (containing the original prompts):

\begin{lstlisting}[language=Python, basicstyle=\ttfamily\small, frame=single, breaklines=true, keywordstyle=\color{blue}\bfseries, stringstyle=\color{green!60!black}, commentstyle=\color{gray}\itshape, showstringspaces=false]
messages = [
    {'role': 'user', 'content': problem}, # raw prompt
    {'role': 'assistant', 'content': base_solution}, # from Phase 1
    {'role': 'user', 'content': refinement_INSTRUCTION}
]
# messages is then packaged into pending_refinement, then passed to run_step() for Phase 2
\end{lstlisting}

with the refinement instruction as: 
\begin{tcolorbox}[colback=gray!5, colframe=gray!50, boxrule=0.5pt,
  left=6pt, right=6pt, top=4pt, bottom=4pt]
\small
Follow this instruction, carefully review your previous solution: \\
1. Go through each calculation step-by-step. Check if there are any errors in calculations, logic, or problem understanding.\\
2. If you find any mistakes, explicitly point out what was wrong and explain the correct approach.\\
3. If the solution is already correct, verify each step and explain it more clearly.\\
4. Finally, after finishing the review, provide your refined solution and answer.
\end{tcolorbox}

The refinement instruction is task-agnostic and contains no correctness signals, ensuring the model learns self-refinement without external supervision.

\subsubsection{Hyperparameter Configuration}

Table~\ref{tab:hparams} summarizes the key hyperparameters for \modelname{} training. 

\begin{table}[t]
\centering
\small
\begin{tabular}{@{}ll@{}}
\toprule
\textbf{Category} & \textbf{Hyperparameter} \\
\midrule
\multicolumn{2}{l}{\textit{GRPO Training}} \\
\midrule
Learning rate & $1 \times 10^{-6}$ \\
PPO clip ratio & $0.2$ \\
Max response length & $3000$ \\
Train batch size & $32$ \\
PPO mini batch size & $8$ \\
Group size ($G$) & $8$ \\
Entropy coefficient & $0.0$ \\
KL penalty in reward & Disabled \\
\midrule
\multicolumn{2}{l}{\textit{Refinement Training}} \\
\midrule
Refinement steps & $2$ (one reasoning + one refinement) \\
Refinement selection mode & random \\
\midrule
\multicolumn{2}{l}{\textit{Generation}} \\
\midrule
Temperature & $1.0$ (train), $0.0$ (val) \\
Top-$p$ & $1.0$ \\
Top-$k$ & $-1$ \\
Max model length & $8192$ \\
\bottomrule
\end{tabular}
\caption{Key hyperparameters for \modelname{} training.}
\label{tab:hparams}
\end{table}

\subsection{Implementation of Training-free Refinement Baselines}
\label{app:baseline_impl}

We compare our trained models against four non-training test-time baselines: \emph{base reasoning}, \emph{one-step refinement}, and two iterative methods inspired by \textsc{Self-Refine}~\citep{madaan2023self} and \textsc{Reflexion}~\citep{shinn2023reflexion}. All baselines are evaluated in \emph{non-thinking} mode. To isolate the effect of the inference procedure from prompt engineering, we keep the refinement instruction fixed across all refinement-based baselines.

\paragraph{Prompt formatting.}
Each evaluation example is rendered with the model's chat template and an added generation prompt. For \textsc{OLMo-3-7B-Instruct}, the tokenizer template automatically prepends a default system message. In the \emph{no-system} ablation, we remove only this automatically injected prefix before decoding, so that the model is evaluated without the template-provided system instruction. For one-step refinement, \textsc{Self-Refine}, and the memory-generation stage of \textsc{Reflexion}, we use the same refinement prompt as shown in Appendix~\ref{app:imp_details}.





\paragraph{base reasoning.}
For each problem \(x_i\), we sample \(n=32\) candidate solutions
\[
y_{i,1}^{(0)}, y_{i,2}^{(0)}, \ldots, y_{i,n}^{(0)}
\]
from the model using temperature \(0.7\), top-\(p=0.8\), top-\(k=20\), and a maximum generation length of 3000 tokens. All second-stage generations (refinement, refinement, and retry) are decoded greedily. Each initial sample defines one \emph{branch}. Iterative updates occur only \emph{within} a branch, and only the final answer of that branch is scored. Consequently, iterative baselines do not receive extra \(\mathrm{pass@}k\) opportunities beyond the original \(n\) branches.

\paragraph{One-step refinement.}
Given a sampled solution \(y_{i,j}^{(0)}\), we construct a second-round prompt containing the original problem, the sampled answer as an assistant turn, and the shared refinement instruction. We then decode a single refined solution \(\hat{y}_{i,j}\) greedily. The set
\[
\{\hat{y}_{i,j}\}_{j=1}^{n}
\]
constitutes the refinement baseline for problem \(i\).

\paragraph{\textsc{Self-Refine}.}
We implement \textsc{Self-Refine} as repeated refinement-and-refinement within the same branch, while keeping the refinement instruction identical to the one-step refinement baseline. Starting from \(y_{i,j}^{(0)}\), we iteratively apply
\[
y_{i,j}^{(t+1)} = \mathcal{M}\!\left(x_i, y_{i,j}^{(t)}, r\right),
\qquad t=0,\dots,T-1,
\]
where \(\mathcal{M}\) denotes the chat model and \(r\) denotes the fixed refinement instruction. We score only the final output \(y_{i,j}^{(T)}\). Under this controlled prompt design, the case \(T=1\) is identical to one-step refinement; therefore, when reporting \textsc{Self-Refine} as a distinct baseline, we use \(T \ge 2\).

\paragraph{\textsc{Reflexion}.}
For \textsc{Reflexion}, we again begin from a base sample \(y_{i,j}^{(0)}\), but use the refinement stage to produce a short verbal memory rather than a final revised answer. Let \(m_{i,j}^{(1)}\) denote the first refinement note. The model then retries the original problem from scratch while conditioning on the accumulated memory:
\[
\tilde{y}_{i,j}^{(t+1)} = \mathcal{M}\!\left(x_i, m_{i,j}^{(\le t)}\right).
\]
We report the final retry output after the prescribed number of retries. In our default setting, we use one retry so that the test-time cost remains comparable to one-step refinement and two-step \textsc{Self-Refine}. Ground-truth answers are never revealed during inference; they are used only for offline evaluation.

\paragraph{Evaluation Protocol.}
A branch is marked correct if its final answer matches the ground-truth solution under the same boxed-answer grader used throughout evaluation. Let \(c_i\) be the number of correct branch finals among the \(n=32\) branches for problem \(i\). Similarly, we estimate \(\mathrm{pass@}k\) following~\citet{chen2021evaluating}. This estimator allows all reported values of \(k\) to be obtained from a single set of sampled generations, without rerunning inference separately for each \(k\).



\paragraph{Reproducibility note.}
All baselines use the same random seed and the same underlying evaluation set. The only difference between methods is how each branch is post-processed at test time: no revision for base reasoning, one revision for refinement, repeated revision for \textsc{Self-Refine}, and retry conditioned on verbal memory for \textsc{Reflexion}.


\section{Additional Results}
\label{app:additional}

\begin{figure*}[h]
    \centering
    \includegraphics[width=1.0\linewidth]{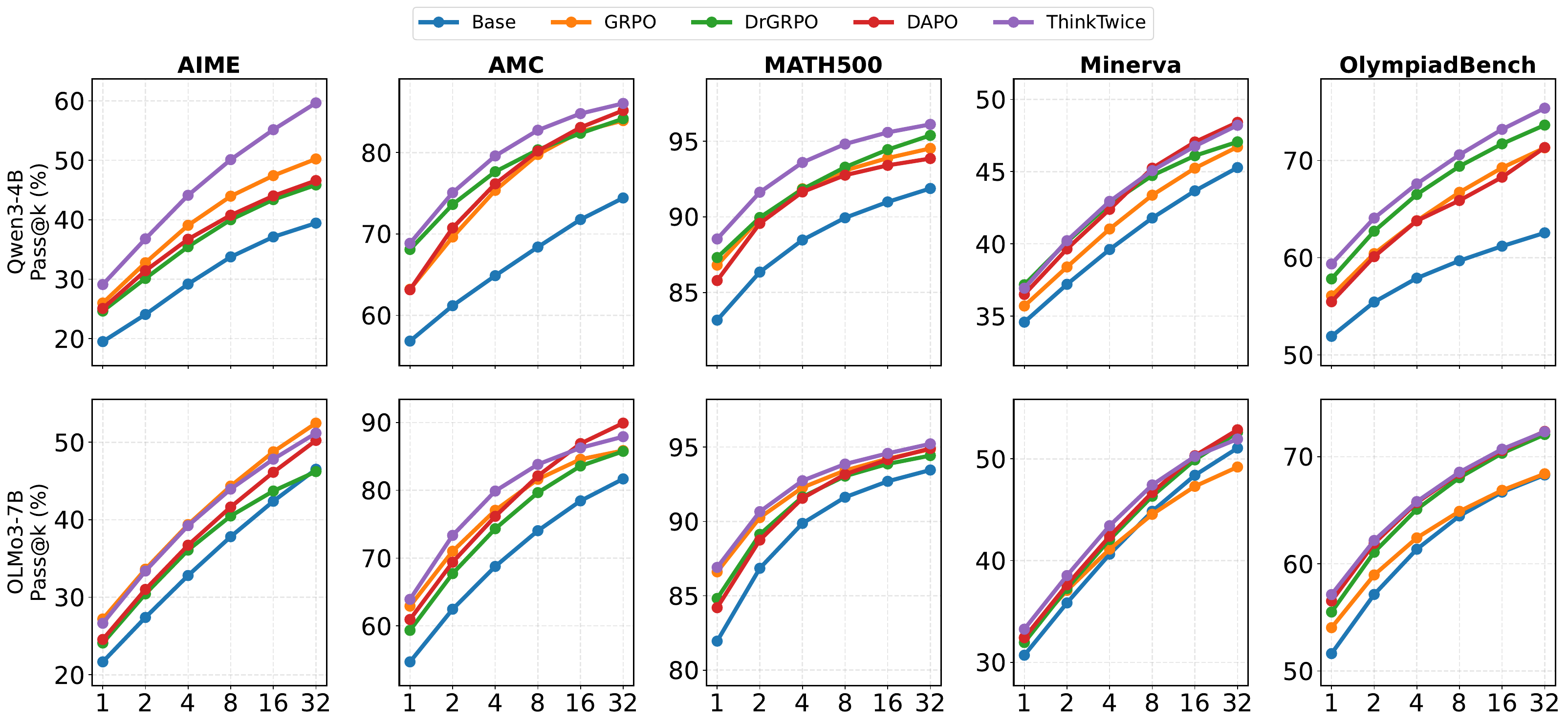}
    \caption{Reasoning pass@$k$ curves across five mathematical reasoning benchmarks for Qwen3-4B (top) and OLMo3-7B (bottom).}
    \label{fig:passatk_base}
\end{figure*}

\begin{figure*}[h]
    \centering
    \includegraphics[width=1.0\linewidth]{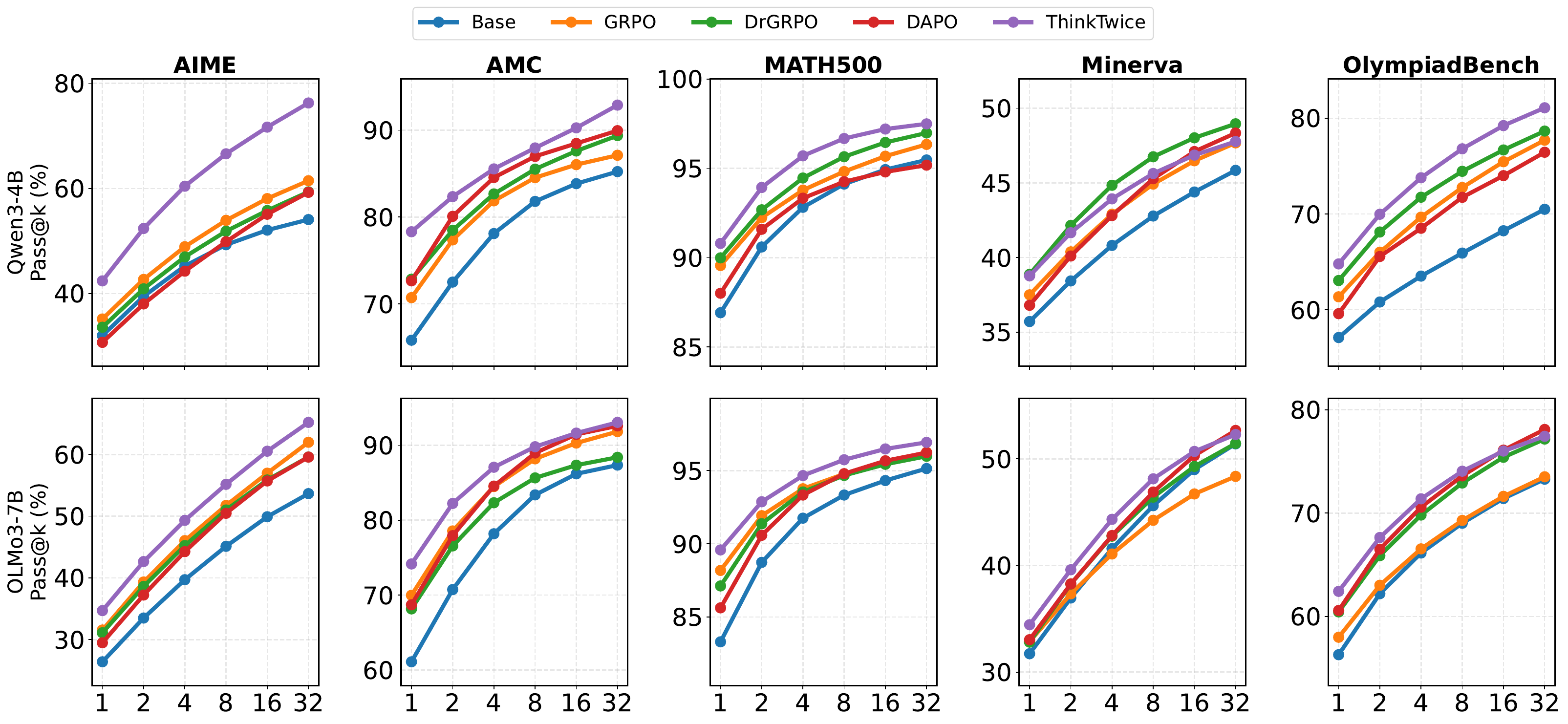}
    \caption{Self-refinement pass@$k$ curves across five mathematical reasoning benchmarks for Qwen3-4B (top) and OLMo3-7B (bottom).}
    \label{fig:passatk_reflection}
\end{figure*}

We provide the full pass@$k$ curves for both reasoning and self-refinement evaluation, complementing the pass@4 results reported in Tables~\ref{tab:reasoning_results} and~\ref{tab:refinement_results}. Figure~\ref{fig:passatk_base} shows reasoning performance and Figure~\ref{fig:passatk_reflection} shows self-refinement performance, both across $k \in \{1, 2, 4, 8, 16, 32\}$ for all five benchmarks and both model families.

For reasoning (Figure~\ref{fig:passatk_base}), ThinkTwice outperforms across nearly all values of $k$ on both Qwen3-4B and OLMo3-7B than other baselines, with its advantage most pronounced on AIME, where the gap widens steadily as $k$ increases. For self-refinement (Figure~\ref{fig:passatk_reflection}), the separation is even more striking: ThinkTwice maintains a clear lead on AIME across the entire $k$ range, reaching nearly 77\% pass@32 on Qwen3-4B compared to roughly 62\% for the next-best baseline. Across both models, the curves confirm that ThinkTwice's improvements are not artifacts of a particular $k$ value but hold consistently across the pass@$k$ spectrum.

\section{Case Studies}
\label{app:case_studies}

The three recurring mechanisms we wanted to document are: (i) \textbf{route switching}, where refinement abandons a bad search path and finds a better method; (ii) \textbf{solution completion}, where the base trace is promising but unfinished and refinement closes the argument; and (iii) \textbf{late-stage exploitation}, where the base trace is already correct but refinement removes exploratory clutter, normalizes the final answer, and presents a shorter proof. Table~\ref{tab:appendix-case-map} summarizes the selected cases.

\begin{table}[h]
\centering
\small
\begin{tabularx}{\linewidth}{@{}l c c X@{}}
\toprule
\textbf{Case} & \textbf{Sample} & \textbf{Checkpoints} & \textbf{Why it matters} \\
\midrule
A & 481 & 100, 240 & Same problem shows both phases of the \ story: early refinement discovers the missing telescoping structure; later refinement gives a much shorter version of the now-correct base proof. \\
B & 497 & 110, 250 & refinement completes a stalled geometric derivation early, then later preserves the method while converting a long decimal-heavy trace into a shorter exact solution. \\
C & 194 & 30, 220 & Early refinement repairs a concrete derivative mistake; later refinement turns a wandering but correct proof into a short direct argument. \\
D & 425 & 100, 220 & Geometry case: early refinement recomputes faulty vector algebra; later refinement keeps the invariant but removes most of the symbolic overhead. \\
\bottomrule
\end{tabularx}
\caption{Case-study map for the detailed qualitative appendix.}
\label{tab:appendix-case-map}
\end{table}

\clearpage
\subsection{Case A: sample 481 --- early structural discovery, late exploitation on the \emph{same} problem}
\label{app:case654}

\begin{problembox}
\textbf{Problem (MATH500, level 5, sample 481).} Let $n$ be a positive integer. Simplify
\[
\frac{(2^4 + \frac{1}{4})(4^4 + \frac{1}{4}) \dotsm [(2n)^4 + \frac{1}{4}]}{(1^4 + \frac{1}{4})(3^4 + \frac{1}{4}) \dotsm [(2n - 1)^4 + \frac{1}{4}]}
\]
\textbf{Ground truth:} $\boxed{8n^2 + 4n + 1}$.

\medskip
\textbf{Why this case is useful.} This is the clearest single-problem narrative we found. At step 100 the \modelname{} reasoning trace is wrong; the refinement keeps the same factorization but discovers the missing telescoping relation and solves the problem. By step 240 the \emph{same} problem has become a late-stage exploitation example: the \modelname{} reasoning is now correct but still very long, while the refinement keeps the proof and compresses it aggressively. The pure-GRPO baseline is informative at both checkpoints: it reaches the right refinement answer at step 100 only after a longer pattern-checking route, and by step 240 its base trace is still wrong while its refinement remains much longer than \modelname{}'s.
\end{problembox}

\begin{table}[h]
\centering
\small
\begin{tabularx}{\linewidth}{@{}l l X X@{}}
\toprule
\textbf{Checkpoint} & \textbf{Run} & \textbf{Base output} & \textbf{refinement output} \\
\midrule
\multirow{2}{*}{Step 100} 
& \modelname{} & \wrong; 1222 words; last boxed value $\boxed{85}$ & \correct; 686 words; $\boxed{8n^2 + 4n + 1}$ \\
& Pure GRPO & \wrong; 1308 words; no final boxed general form & \correct; 892 words; $\boxed{8n^2 + 4n + 1}$ \\
\midrule
\multirow{2}{*}{Step 240}
& \modelname{} & \correct; 1222 words; $\boxed{8n^2 + 4n + 1}$ & \correct; 358 words; $\boxed{8n^2 + 4n + 1}$ \\
& Pure GRPO & \wrong; 1243 words; no final boxed general form & \correct; 1067 words; $\boxed{8n^2 + 4n + 1}$ \\
\bottomrule
\end{tabularx}
\caption{Case A summary. The same problem exhibits both early correction and late-stage compression.}
\label{tab:case654-summary}
\end{table}

\subsubsection*{Early checkpoint (step 100): refinement finds the missing telescoping argument}
The \modelname{} reasoning answer already has the right local algebraic move,
\[
x^4 + \tfrac14 = \left(x^2 - x + \tfrac12\right)\left(x^2 + x + \tfrac12\right),
\]
but it never turns that factorization into a general product formula. Instead, the trace drifts into pattern-matching on small values of $n$:

\begin{excerptbox}{\modelname{} reasoning, step 100 (wrong; 1222 words)}
``But maybe we can look for a telescoping pattern? Let's compute the expression for small values of $n$ and see if we can spot a pattern.''
It checks $n=1,2,3$ and ends with the last computed value $\boxed{85}$, which is the $n=3$ instance rather than a general closed form.
\end{excerptbox}

The refinement does \emph{not} restart from scratch. Instead, it keeps the factorization and adds the structural observation that the base trace missed. After simplifying a single term to
\[
\frac{8k^2 + 4k + 1}{8k^2 - 12k + 5},
\]
it makes the crucial recurrence explicit:

\begin{excerptbox}{\modelname{} refinement, step 100 (correct; 686 words)}
``So the denominator of term $k$ is equal to the numerator of term $k-1$. Therefore, the entire product telescopes.''
It then rewrites the product as
\[
\prod_{k=1}^{n} \frac{b_{k+1}}{b_k} = \frac{b_{n+1}}{b_1}, \qquad b_k = 8k^2 - 12k + 5,
\]
and concludes with $\boxed{8n^2 + 4n + 1}$.
\end{excerptbox}

The same-step pure-GRPO baseline refinement also reaches the right answer, but by a looser and longer route. It explicitly computes several small cases before inferring the formula:

\begin{excerptbox}{Pure-GRPO refinement, step 100 (correct; 892 words)}
``Step 4: Look for telescoping pattern. We now compute this product for small $n$ to detect a pattern.''
Later it writes ``Perfect! So the entire product simplifies to $8n^2 + 4n + 1$.''
\end{excerptbox}

The important qualitative difference here is not that \modelname{} is the only system that can solve the problem at this checkpoint; rather, \modelname{}'s refinement is already more \emph{structural}. The base gets trapped in empirical pattern-chasing, while the refinement explicitly identifies the recurrence that makes telescoping work.

\subsubsection*{Late checkpoint (step 240): same proof idea, much shorter refinement}
By step 240 the \modelname{} reasoning trace has learned the right proof family: it now factors correctly and even states the recurrence
\[
a_k = b_{k+1}
\]
for the telescoping terms. But the base answer is still long and exploratory, with repeated recalculations and explicit small-$n$ checks. A representative moment is:

\begin{excerptbox}{\modelname{} reasoning, step 240 (correct; 1222 words)}
``Wait, let's do it carefully.''
Later, after reaching the closed form, it still spends another long block numerically checking $n=1$ and $n=2$ from scratch.
\end{excerptbox}

The refinement compresses that same proof into a direct $b_{k+1}/b_k$ argument:

\begin{excerptbox}{\modelname{} refinement, step 240 (correct; 358 words)}
It defines
\[
b_k = 8k^2 - 12k + 5,
\]
computes
\[
b_{k+1} = 8k^2 + 4k + 1,
\]
and then writes
\[
\prod_{k=1}^n \frac{b_{k+1}}{b_k} = \frac{b_{n+1}}{b_1} = 8n^2 + 4n + 1.
\]
The answer is then immediately boxed as $\boxed{8n^2 + 4n + 1}$.
\end{excerptbox}

The same-step pure-GRPO baseline remains less stable. Its base answer is still wrong, and even its correct refinement is over three times longer than \modelname{}'s and continues to rely on guess-and-check language:

\begin{excerptbox}{Pure-GRPO refinement, step 240 (correct; 1067 words)}
After deriving the same single-term ratio, it still pauses for several empirical checks: ``Wait: $E_1 = 13$ ... $E_2 = 41$ ... $E_3 = 85$? Let's verify.'' Only later does it box $8n^2 + 4n + 1$.
\end{excerptbox}

This case is the cleanest qualitative support for the ``explore early, exploit later'' story. Early on, refinement discovers the missing structure that the base answer cannot yet use. Later on, once the base has learned the right route, refinement becomes the short and decisive version of that same route.

\subsection{Case B: sample 497 --- completing a stalled geometric derivation, then normalizing the final answer}
\label{app:case670}

\begin{problembox}
\textbf{Problem (MATH500, level 5, sample 497).} An equiangular octagon has four sides of length $1$ and four sides of length $\frac{\sqrt{2}}{2}$, arranged so that no two consecutive sides have the same length. What is the area of the octagon?

\textbf{Ground truth:} $\boxed{\frac{7}{2}}$.

\medskip
\textbf{Why this case is useful.} This is our clearest ``finish the job'' geometry example. At step 110 the \modelname{} reasoning trace has the right global setup --- directions, vectors, cumulative coordinates, and shoelace --- but it stalls in the area computation and never produces a final boxed answer. The refinement keeps the same coordinate plan and makes it executable. By step 250 the base trace is correct, but it is still long and reports the area as the decimal $3.5$; the refinement keeps the same proof while shortening it and normalizing the answer to the exact form $\frac{7}{2}$.
\end{problembox}

\begin{table}[h]
\centering
\small
\begin{tabularx}{\linewidth}{@{}l l X X@{}}
\toprule
\textbf{Checkpoint} & \textbf{Run} & \textbf{Base output} & \textbf{refinement output} \\
\midrule
\multirow{2}{*}{Step 110} 
& \modelname{} & \wrong; 1280 words; no final boxed answer & \correct; 633 words; $\boxed{\frac{7}{2}}$ \\
& Pure GRPO & \wrong; 1284 words; no final boxed answer & \wrong; 970 words; no final boxed answer \\
\midrule
\multirow{2}{*}{Step 250}
& \modelname{} & \correct; 1303 words; $\boxed{3.5}$ & \correct; 591 words; $\boxed{\frac{7}{2}}$ \\
& Pure GRPO & \correct; 1209 words; $\boxed{\frac{7}{2}}$ & \correct; 884 words; $\boxed{\frac{7}{2}}$ \\
\bottomrule
\end{tabularx}
\caption{Case B summary. Early refinement completes the coordinate-shoelace derivation; late refinement keeps the method but presents it more cleanly.}
\label{tab:case670-summary}
\end{table}

\subsubsection*{Early checkpoint (step 110): the base has the right plan but cannot close it}
The \modelname{} reasoning answer chooses a sensible geometry strategy: it sets the side directions to $0^\circ,45^\circ,\ldots,315^\circ$, alternates the side lengths, constructs the cumulative coordinates, and then moves to the shoelace formula. The failure is not conceptual; it is procedural. The trace keeps changing how it wants to compute the area:

\begin{excerptbox}{\modelname{} reasoning, step 110 (wrong; 1280 words)}
The trace reaches the area stage and then hesitates: ``We can compute the area of a polygon given its vertices using the shoelace formula. Wait --- actually, that's not quite right. Better: use the vector cross-product method.'' It never reaches a final boxed answer.
\end{excerptbox}

The refinement keeps the same coordinate walk but turns it into a checkable proof. It explicitly verifies the alternating side assignment, the vector directions, the cumulative coordinates, and finally the shoelace sum:

\begin{excerptbox}{\modelname{} refinement, step 110 (correct; 633 words)}
``Side 1: length 1, $0^\circ$ \(\to (1,0)\) ... Side 8: length $\frac{\sqrt{2}}{2}$, $315^\circ$ \(\to (\tfrac12,-\tfrac12)\).''
After listing the vertices, it computes the shoelace terms and concludes with
\[
\boxed{\frac{7}{2}}.
\]
\end{excerptbox}

The same-step pure-GRPO baseline is informative because it fails in exactly the place where \modelname{}'s refinement succeeds. The baseline refinement is still expanding shoelace terms at the output limit and never boxes a final answer:

\begin{excerptbox}{Pure-GRPO refinement, step 110 (wrong; 970 words)}
The last available lines are still inside the term-by-term expansion,
\[
 x_6 y_7 - x_7 y_6 = \left(-\frac{\sqrt{2}}{2}\right)\left(\frac{\sqrt{2}}{2}\right) - \left(-\frac{\sqrt{2}}{2}\right)\left(1+\frac{\sqrt{2}}{2}\right),
\]
with no final boxed answer.
\end{excerptbox}

This is a useful mechanism example because refinement is not inventing a completely different strategy. The base trace already knows the right global method; refinement makes it \emph{executable}, removes the indecision at the area step, and finishes the derivation.

\subsubsection*{Late checkpoint (step 250): same geometry, cleaner ending, exact final form}
By step 250 the \modelname{} reasoning answer solves the problem, but the trace is still long and numerically heavy. It ends with the decimal result
\[
\boxed{3.5},
\]
after a lengthy shoelace calculation. The last few lines read like a running calculator tape:

\begin{excerptbox}{\modelname{} reasoning, step 250 (correct; 1303 words)}
``$0 + 0.5 + 1.5 + 1.5 + 2.0 + 1.0 + 0.5 + 0 = 7.0$. So total sum = $7.0$. Area = $\frac12 \times |7.0| = \boxed{3.5}$.''
\end{excerptbox}

The refinement keeps the same coordinate proof but tightens the presentation and restores the exact answer format:

\begin{excerptbox}{\modelname{} refinement, step 250 (correct; 591 words)}
It verifies each side vector once, summarizes the shoelace sum compactly, and then states
\[
\boxed{\frac{7}{2}}.
\]
The refinement keeps the proof exact instead of finishing in decimal form.
\end{excerptbox}

The pure-GRPO baseline is also correct at this late checkpoint, but its refinement is still substantially longer (884 words versus 591) and includes an extra ``reasonableness check'' section after the exact shoelace computation. This is the kind of late-stage difference that is easy to miss if one looks only at accuracy: both systems are correct, but \modelname{}'s refinement is much closer to the polished proof style we would actually want to show a user.

\subsection{Case C: sample 194 --- concrete error repair early, concise proof late}
\label{app:case367}

\begin{problembox}
\textbf{Problem (MATH500, level 5, sample 194).} Find the maximum value of
\[
\frac{x-y}{x^4+y^4+6}
\]
over all real numbers $x$ and $y$.

\textbf{Ground truth:} $\boxed{\frac{1}{4}}$.

\medskip
\textbf{Why this case is useful.} This case separates two different refinement behaviors. At step 30 the base trace reaches the right one-variable reduction but makes a concrete derivative arithmetic error; the refinement explicitly identifies and fixes that error. By step 220 the base trace is correct, yet still wanders through multiple abandoned ideas before it settles on the final proof. The refinement turns that wandering derivation into a much shorter argument. The same-step pure-GRPO baseline is already correct at both checkpoints, so this case is \emph{not} about a baseline failure; it is included because it shows that \modelname{} learns both concrete bug repair and later proof compression on the same problem.
\end{problembox}

\begin{table}[h]
\centering
\small
\begin{tabularx}{\linewidth}{@{}l l X X@{}}
\toprule
\textbf{Checkpoint} & \textbf{Run} & \textbf{Base output} & \textbf{refinement output} \\
\midrule
\multirow{2}{*}{Step 30} 
& \modelname{} & \wrong; 1204 words; no final boxed answer & \correct; 719 words; $\boxed{\frac{1}{4}}$ \\
& Pure GRPO & \correct; 1236 words; $\boxed{\frac{1}{4}}$ & \correct; 1181 words; $\boxed{\frac{1}{4}}$ \\
\midrule
\multirow{2}{*}{Step 220}
& \modelname{} & \correct; 1373 words; $\boxed{\frac{1}{4}}$ & \correct; 374 words; $\boxed{\frac{1}{4}}$ \\
& Pure GRPO & \correct; 906 words; $\boxed{\frac{1}{4}}$ & \correct; 755 words; $\boxed{\frac{1}{4}}$ \\
\bottomrule
\end{tabularx}
\caption{Case C summary. Early refinement fixes a concrete derivative mistake; late refinement turns a wandering proof into a short direct argument.}
\label{tab:case367-summary}
\end{table}

\subsubsection*{Early checkpoint (step 30): explicit derivative bug repair}
The \modelname{} reasoning answer makes a strong start. It correctly reduces the problem to maximizing
\[
f(d)=\frac{d}{\tfrac18 d^4 + 6}
\]
by minimizing $x^4+y^4$ at fixed difference $d=x-y$. The failure is a concrete arithmetic error in the derivative step:

\begin{excerptbox}{\modelname{} reasoning, step 30 (wrong; 1204 words)}
After deriving
\[
f'(d)=\frac{-\tfrac38 d^4 + 6}{(\tfrac18 d^4 + 6)^2},
\]
it sets the numerator to zero and writes
\[
-\tfrac38 d^4 + 6 = 0 \Rightarrow d^4 = 1,
\]
which is wrong; the correct value is $d^4=16$.
\end{excerptbox}

The refinement is unusually explicit about what went wrong:

\begin{excerptbox}{\modelname{} refinement, step 30 (correct; 719 words)}
``Mistake in previous step: I incorrectly computed $d^4 = 1$ instead of $d^4 = 16$.''
It then corrects the derivative equation to
\[
-\tfrac38 d^4 + 6 = 0 \Rightarrow d^4 = 16 \Rightarrow d=2,
\]
substitutes $x=1$, $y=-1$, and concludes with $\boxed{\frac14}$.
\end{excerptbox}

The pure-GRPO baseline already solves this problem at the same early checkpoint, so the point of this example is narrower: \modelname{}'s refinement learns a genuine \emph{bug-fixing} behavior, not just a generic ``rewrite more clearly'' behavior.

\subsubsection*{Late checkpoint (step 220): preserving correctness while removing exploratory clutter}
At step 220 the \modelname{} reasoning answer is correct, but it is still visibly exploratory. It tries multiple proof routes, including one that it explicitly rejects:

\begin{excerptbox}{\modelname{} reasoning, step 220 (correct; 1373 words)}
The trace first tries an inequality route,
\[
z^4 + 1 \ge 4z \quad \text{for all real } z,
\]
then immediately checks it at $z=1$ and notes ``$1-4+1=-2$ --- not non-negative, so that inequality fails.'' Only after this detour does it return to the correct calculus proof.
\end{excerptbox}

The refinement starts much closer to the final proof. It opens with the extremal point, then proves optimality directly:

\begin{excerptbox}{\modelname{} refinement, step 220 (correct; 374 words)}
``We try a substitution: let $x=1$, $y=-1$.''
It then proves
\[
\frac{x-y}{x^4+y^4+6} \le \frac14
\]
by minimizing
\[
f(x,y)=x^4+y^4-4x+4y+6,
\]
checks the critical point $(1,-1)$, and ends with $\boxed{\frac14}$.
\end{excerptbox}

The same-step pure-GRPO baseline is also correct, but its refinement is more exploratory and almost twice as long. It begins with several test points (e.g., $x=1,y=0$; $x=1,y=-1$; $x=2,y=-1$) before moving to the formal proof. This makes the late-stage contrast very clear: once the base policy has learned the right answer, \modelname{}'s refinement behaves like an exploitation layer that strips away the dead ends.

\subsection{Case D: sample 425 --- vector recomputation as a backup geometry case}
\label{app:case598}

\begin{problembox}
\textbf{Problem (MATH500, level 5, sample 425).} Let $G$ and $H$ denote the centroid and orthocenter of triangle $ABC$, respectively. Let $F$ be the midpoint of $\overline{GH}$. Express $AF^2 + BF^2 + CF^2$ in terms of the side lengths $a,b,c$ and circumradius $R$.

\textbf{Ground truth:} $\boxed{3R^2}$.

\medskip
\textbf{Why this case is useful.} This is a strong backup geometry example. Early on, the \modelname{} reasoning trace carries out the right vector setup but leaves the derivation in an unresolved mixed term involving $\vec S \cdot \vec O$. The refinement recomputes the vector expansion carefully and substitutes the missing circumcenter identity. By step 220 the base trace has learned the cleaner circumcenter-at-origin derivation, and the refinement compresses it further.
\end{problembox}

\begin{table}[h]
\centering
\small
\begin{tabularx}{\linewidth}{@{}l l X X@{}}
\toprule
\textbf{Checkpoint} & \textbf{Run} & \textbf{Base output} & \textbf{refinement output} \\
\midrule
\multirow{2}{*}{Step 100} 
& \modelname{} & \wrong; 1138 words; no final boxed answer & \correct; 795 words; $\boxed{3R^2}$ \\
& Pure GRPO & \correct; 711 words; $\boxed{3R^2}$ & \correct; 1046 words; $\boxed{3R^2}$ \\
\midrule
\multirow{2}{*}{Step 220}
& \modelname{} & \correct; 746 words; $\boxed{3R^2}$ & \correct; 410 words; $\boxed{3R^2}$ \\
& Pure GRPO & \correct; 1259 words; $\boxed{3R^2}$ & \correct; 993 words; $\boxed{3R^2}$ \\
\bottomrule
\end{tabularx}
\caption{Case D summary. A backup geometry example showing early recomputation and late compression.}
\label{tab:case598-summary}
\end{table}

\subsubsection*{Early checkpoint (step 100): refinement repairs the vector algebra}
The \modelname{} reasoning answer gets close to the goal but stops at an unresolved mixed term:

\begin{excerptbox}{\modelname{} reasoning, step 100 (wrong; 1138 words)}
After expanding the squared norms, the base trace reduces the expression to
\[
AF^2 + BF^2 + CF^2 = -2(\vec S \cdot \vec O) + 3\lVert \vec O \rVert^2 + \sum \lVert \vec X \rVert^2,
\]
but never substitutes the identity needed to eliminate the remaining $\vec S \cdot \vec O$ term.
\end{excerptbox}

The refinement explicitly decides to recompute instead of trusting the previous algebra:

\begin{excerptbox}{\modelname{} refinement, step 100 (correct; 795 words)}
``Let's recompute the expression carefully, term by term.''
It recomputes $\vec S \cdot \vec H$ and $\lVert \vec H \rVert^2$, then uses
\[
\sum_{X=A,B,C} \left( \|\vec X\|^2 - 2\vec X\cdot\vec O + \|\vec O\|^2 \right) = 3R^2
\]
to replace the mixed term and conclude with $\boxed{3R^2}$.
\end{excerptbox}

The pure-GRPO baseline is also correct here, but its refinement is substantially longer and contains its own visible false starts (including the line ``Still wrong'' while checking an auxiliary identity). This makes the \modelname{} refinement look less like another exploratory search and more like a controlled repair.

\subsubsection*{Late checkpoint (step 220): concise invariant-based proof}
By step 220 the \modelname{} reasoning answer has already adopted the cleaner circumcenter-at-origin derivation. The refinement compresses it to a short invariant argument:

\begin{excerptbox}{\modelname{} refinement, step 220 (correct; 410 words)}
It places the circumcenter at the origin, uses
\[
\vec H = \vec A + \vec B + \vec C, \qquad \vec G = \tfrac13(\vec A + \vec B + \vec C), \qquad \vec F = \tfrac23(\vec A + \vec B + \vec C),
\]
and then cancels the cross terms in
\[
\sum AF^2 = 3\|\vec F\|^2 + \sum \|\vec A\|^2 - 2\vec F\cdot(\vec A+\vec B+\vec C)
\]
to get $\boxed{3R^2}$.
\end{excerptbox}

The pure-GRPO baseline is also correct at this checkpoint, but both its base and refinement are much longer (1259 and 993 words). This is a good backup example when we want a geometry case that is less about a wrong final answer and more about the shift from symbolic overhead to a clean invariant.

\subsection{Key takeaway from the case study}
Across all four examples, the same qualitative pattern recurs. Early in training, refinement is most useful because it can \emph{change what the model is doing}: it can discover a missing structural relation (Case A), finish a derivation that the base trace cannot close (Case B), or repair a concrete algebraic error (Case C). Later in training, once the base policy more often lands on the right route, refinement shifts toward \emph{exploitation}: it preserves correctness while removing numerical detours, dead-end branches, and unnecessary verification blocks (Cases A--D). This is exactly the qualitative behavior we would expect if joint reasoning-and-refinement training is teaching a reusable refinement policy rather than merely adding another pass of unconstrained generation.

\end{document}